%% file: main.tex
\documentclass[preprint, review, 12pt]{elsarticle}
\usepackage{enumitem}
\usepackage{lineno,hyperref}
\modulolinenumbers[5]
\usepackage{floatpag}

\journal{Energy and AI}

\usepackage[acronym,nomain]{glossaries}
\makenoidxglossaries
\newacronym{oat}{OAT}{outside-air-temperature}
\newacronym{es}{ES}{Energy Signature}
\newacronym{mv}{M\&V}{measurement and verification}
\newacronym{svr}{SVR}{support-vector regression}

\usepackage{amsthm}% for \@addpunct
\makeatletter
\def\els@aparagraph[#1]#2{\elsparagraph[#1]{#2\@addpunct{.}}}
\def\els@bparagraph#1{\elsparagraph*{#1\@addpunct{.}}}
\makeatother
\usepackage{booktabs}
\usepackage{adjustbox}
\usepackage[font=small,labelfont=bf]{caption}
\usepackage{floatrow}
\newcommand{\abbreviations}[1]{%
  \nonumnote{\textit{Abbreviations:\enspace}#1}}
\usepackage{amssymb}

\begin{document}

\begin{frontmatter}

\title{Using Bayesian deep learning approaches for uncertainty-aware building energy surrogate models}

% Group authors per affiliation:
\author[1]{Paul Westermann\corref{mycorrespondingauthor}}\ead{pwestermann@uvic.ca}
\author[1]{Ralph Evins}\ead{revins@uvic.ca}
\address[1]{Energy and Cities Group\\ Department of Civil Engineering\\ University of Victoria, Canada}
\cortext[mycorrespondingauthor]{Corresponding Author}



%\ead{pwestermann@uvic.ca}

\begin{abstract}
Fast machine learning-based surrogate models are trained to emulate slow, high-fidelity engineering simulation models to accelerate engineering design tasks. This introduces uncertainty as the surrogate is only an approximation of the original model.\\
Bayesian methods can quantify that uncertainty, and deep learning models exist that follow the Bayesian paradigm. These models, namely Bayesian neural networks and Gaussian process models, enable us to give predictions together with an estimate of the model's uncertainty. As a result we can derive uncertainty-aware surrogate models that can automatically suspect unseen design samples to cause large emulation errors. For these samples the high-fidelity model can be queried instead. This outlines how the Bayesian paradigm allows us to hybridize fast, but approximate, and slow, but accurate models. \\
In this paper, we train two types of Bayesian models, dropout neural networks and stochastic variational Gaussian Process models, to emulate a complex high dimensional building energy performance simulation problem. The surrogate model processes 35 building design parameters (inputs) to estimate 12 different performance metrics (outputs). We benchmark both approaches, prove their accuracy to be competitive, and show that errors can be reduced by up to 30\% when the 10\% of samples with the highest uncertainty are transferred to the high-fidelity model.
%Read Wates' paper on aleatoric and epistemic uncertainty.
\end{abstract}

\begin{keyword}
Surrogate modelling \sep metamodel \sep building performance simulation \sep uncertainty \sep Bayesian deep learning \sep Gaussian Process \sep Bayesian neural network
%\MSC[2010] 00-01\sep  99-00 --> what is this?
\end{keyword}
\abbreviations{BDL: Bayesian deep learning; BNN: Bayesian neural network; SVGP: stochastic-variational Gaussian Process; DoE: design-of-experiment; ReLU: rectified linear unit;}

\end{frontmatter}

\section*{Highlights}
\begin{itemize}
\item Training of uncertainty-aware engineering surrogate models.
\item Comparing deep Bayesian neural networks and Gaussian process models.
\item Uncertainty estimates can identify and mitigate errors in surrogate models.
\end{itemize}

%\tableofcontents

\input{01Introduction}

\input{02Background}

\input{03Methodology}
\input{04Results}

\input{05DiscussionConclusion}

\bibliography{references}

\section*{Appendix}
\vspace*{-.6cm}
\begin{table}[h]
\hspace*{-3cm}
\setlength{\tabcolsep}{2pt} 
\renewcommand{\arraystretch}{.5}	
\begin{footnotesize}
\begin{tabular}{lll|ll|ll}
%\toprule
{} &              \multicolumn{2}{c|}{$R^2$} &              \multicolumn{2}{c|}{$MAPE$} &           \multicolumn{2}{c}{$APE90$}  \\
{} &              BNN &     SVGP &           BNN &              SVGP&           BNN &           SVGP \\
\hline
Pumps [MWh]                 &  \textbf{0.990}$ \pm 0.001$ &  $0.983 \pm 0.001$ &  \textbf{7.180}$ \pm 0.180$ &   $8.530 \pm 0.260$ &  \textbf{14.830}$ \pm 0.510$ &  $17.950 \pm 0.610$ \\
Heating supply, Other [MWh] &  \textbf{0.990}$ \pm 0.003$ &  $0.977 \pm 0.001$ &  \textbf{9.820}$ \pm 0.350$ &  $12.490 \pm 0.430$ &  \textbf{22.300}$ \pm 0.750$ &  $29.300 \pm 1.480$ \\
Fans [MWh]                  &  \textbf{0.991}$ \pm 0.004$ &  $0.988 \pm 0.001$ &  $8.630 \pm 0.380$ &   \textbf{8.530}$ \pm 0.250$ &  \textbf{18.120}$ \pm 0.770$ &  $18.280 \pm 0.540$ \\
Heating supply, Elec. [MWh] &  \textbf{0.992}$ \pm 0.001$ &  $0.986 \pm 0.000$ &  \textbf{7.150}$ \pm 0.290$ &   $8.670 \pm 0.360$ &  \textbf{15.130}$ \pm 0.290$ &  $18.260 \pm 0.900$ \\
Heating supply, Gas [MWh]   &  \textbf{0.992}$ \pm 0.002$ &  $0.973 \pm 0.001$ &  \textbf{9.400}$ \pm 0.380$ &  $13.230 \pm 0.220$ &  \textbf{21.440}$ \pm 0.620$ &  $30.480 \pm 0.520$ \\
Cooling supply, Elec. [MWh] &  $0.992 \pm 0.002$ &  \textbf{0.998}$ \pm 0.000$ &  $3.550 \pm 0.200$ &   \textbf{2.820}$ \pm 0.100$ &   $7.490 \pm 0.560$ &   \textbf{5.820}$ \pm 0.200$ \\
Heating demand [MWh]        &  $0.995 \pm 0.001$ &  \textbf{0.996}$ \pm 0.000$ &  $3.960 \pm 0.330$ &   \textbf{3.710}$ \pm 0.080$ &   $8.040 \pm 0.710$ &   \textbf{7.800}$ \pm 0.250$ \\
Cooling demand [MWh]        &  \textbf{0.997}$ \pm 0.000$ &  \textbf{0.997}$ \pm 0.000$ &  $2.440 \pm 0.050$ &   \textbf{2.270}$ \pm 0.060$ &   $4.980 \pm 0.090$ &   \textbf{4.700}$ \pm 0.110$ \\
Interior lights [MWh]       &  $0.998 \pm 0.000$ &  \textbf{0.999}$ \pm 0.000$ &  $2.410 \pm 0.100$ &   \textbf{1.590}$ \pm 0.080$ &   $5.050 \pm 0.160$ &   \textbf{3.150}$ \pm 0.270$ \\
Interior equipment [MWh]    &  \textbf{0.998}$ \pm 0.000$ &  \textbf{0.998}$ \pm 0.000$ &  $2.790 \pm 0.100$ &   \textbf{1.410}$ \pm 0.120$ &   $5.650 \pm 0.200$ &   \textbf{2.600}$ \pm 0.250$ \\
Water heating, Gas [MWh]    &  $0.999 \pm 0.000$ &  \textbf{1.000}$ \pm 0.000$ &  $1.220 \pm 0.130$ &   \textbf{0.250}$ \pm 0.070$ &   $2.590 \pm 0.260$ &   \textbf{0.430}$ \pm 0.090$ \\
PV Generation [MWh]         &  \textbf{0.999}$ \pm 0.000$ &  \textbf{0.999}$ \pm 0.001$ &  $3.030 \pm 0.090$ &   \textbf{1.290}$ \pm 0.090$ &   $6.040 \pm 0.100$ &   \textbf{2.200}$ \pm 0.150$ \\
\bottomrule
\vspace*{-.6cm}
\end{tabular}
\caption{\textbf{Results of the accuracy of the Bayesian models.}}\label{tab:first}
\hspace*{-3cm}
\end{footnotesize}	
\end{table}

\begin{table}

%%%%%%%%%%%%%% Table 1 %%%%%%%%%%%%%%%%%

\begin{flushleft}
\textit{(i) $R^2$-score}
\end{flushleft}
\vspace*{-.5cm}
\setlength{\tabcolsep}{2pt}
\renewcommand{\arraystretch}{.5}
\begin{footnotesize}
\begin{tabular}{lc|ccc}
{} &              ANN &              BNN &            BNN$_{90\%}$ &BNN$_{80\%}$  \\
\midrule
Pumps [MWh]                 &  \textbf{0.992}$ \pm 0.000$ &  $0.990 \pm 0.001$ &  $0.989 \pm 0.001$ &  $0.989 \pm 0.001$ \\
Heating supply, Other [MWh] &  \textbf{0.995}$ \pm 0.001$ &  $0.990 \pm 0.003$ &  $0.989 \pm 0.004$ &  $0.988 \pm 0.004$ \\
Fans [MWh]                  &  \textbf{0.994}$ \pm 0.002$ &  $0.991 \pm 0.004$ &  $0.990 \pm 0.004$ &  $0.989 \pm 0.004$ \\
Heating supply, Elec. [MWh] &  \textbf{0.994}$ \pm 0.000$ &  $0.992 \pm 0.001$ &  $0.992 \pm 0.001$ &  $0.992 \pm 0.001$ \\
Heating supply, Gas [MWh]   &  \textbf{0.995}$ \pm 0.001$ &  $0.992 \pm 0.002$ &  $0.992 \pm 0.002$ &  $0.991 \pm 0.002$ \\
Cooling supply, Elec. [MWh] &  \textbf{0.994}$ \pm 0.001$ &  $0.992 \pm 0.002$ &  $0.993 \pm 0.001$ &  $0.992 \pm 0.002$ \\
Heating demand [MWh]        &  \textbf{0.996}$ \pm 0.000$ &  $0.995 \pm 0.001$ &  $0.995 \pm 0.001$ &  $0.993 \pm 0.002$ \\
Cooling demand [MWh]        &  \textbf{0.997}$ \pm 0.000$ &  $0.997 \pm 0.000$ &  $0.996 \pm 0.000$ &  $0.995 \pm 0.000$ \\
Interior lights [MWh]       &  \textbf{0.999}$ \pm 0.000$ &  $0.998 \pm 0.000$ &  $0.997 \pm 0.000$ &  $0.997 \pm 0.000$ \\
Interior equipment [MWh]    &  \textbf{0.999}$ \pm 0.000$ &  $0.998 \pm 0.000$ &  $0.998 \pm 0.000$ &  $0.997 \pm 0.000$ \\
Water heating, Gas [MWh]    &  \textbf{1.000}$ \pm 0.000$ &  $0.999 \pm 0.000$ &  $0.998 \pm 0.000$ &  $0.998 \pm 0.001$ \\
PV Generation [MWh]         &  \textbf{1.000}$ \pm 0.000$ &  $0.999 \pm 0.000$ &  $0.998 \pm 0.000$ &  $0.998 \pm 0.000$ \\
\bottomrule
\end{tabular}
\end{footnotesize}
%%%%%%%%%%%%%% Table 2 %%%%%%%%%%%%%%%%%

\begin{flushleft}
\textit{(ii) $MAPE$}
\end{flushleft}
\vspace*{-.5cm}
\setlength{\tabcolsep}{2pt} 
\begin{footnotesize}
\begin{tabular}{lc|ccc}
{} &              ANN &              BNN &            BNN$_{90\%}$ &BNN$_{80\%}$  \\
\midrule
Pumps [MWh]                 &  $6.480 \pm 0.170$ &  $7.180 \pm 0.180$ &  $6.200 \pm 0.130$ &  \textbf{5.850}$ \pm 0.130$ \\
Heating supply, Other [MWh] &  $8.550 \pm 0.630$ &  $9.820 \pm 0.350$ &  $8.380 \pm 0.310$ &  \textbf{7.480}$ \pm 0.410$ \\
Fans [MWh]                  &  $7.610 \pm 1.000$ &  $8.630 \pm 0.380$ &  $7.300 \pm 0.470$ &  \textbf{6.690}$ \pm 0.540$ \\
Heating supply, Elec. [MWh] &  $6.530 \pm 0.370$ &  $7.150 \pm 0.290$ &  $6.070 \pm 0.270$ &  \textbf{5.670}$ \pm 0.320$ \\
Heating supply, Gas [MWh]   &  $8.040 \pm 0.220$ &  $9.400 \pm 0.380$ &  $7.880 \pm 0.370$ &  \textbf{7.190}$ \pm 0.400$ \\
Cooling supply, Elec. [MWh] &  $3.280 \pm 0.260$ &  $3.550 \pm 0.200$ &  $3.320 \pm 0.200$ &  \textbf{3.150}$ \pm 0.170$ \\
Heating demand [MWh]        &  $3.710 \pm 0.290$ &  $3.960 \pm 0.330$ &  $3.550 \pm 0.370$ &  \textbf{3.410}$ \pm 0.370$ \\
Cooling demand [MWh]        &  \textbf{2.240}$ \pm 0.160$ &  $2.440 \pm 0.050$ &  $2.310 \pm 0.050$ &  $2.250 \pm 0.060$ \\
Interior lights [MWh]       &  \textbf{1.830}$ \pm 0.170$ &  $2.410 \pm 0.100$ &  $2.290 \pm 0.090$ &  $2.180 \pm 0.070$ \\
Interior equipment [MWh]    &  $2.810 \pm 0.390$ &  $2.790 \pm 0.100$ &  $2.290 \pm 0.080$ &  \textbf{2.130}$ \pm 0.090$ \\
Water heating, Gas [MWh]    &  \textbf{0.660}$ \pm 0.060$ &  $1.220 \pm 0.130$ &  $1.110 \pm 0.130$ &  $1.050 \pm 0.120$ \\
PV Generation [MWh]         &  \textbf{1.650}$ \pm 0.120$ &  $3.030 \pm 0.090$ &  $1.900 \pm 0.150$ &  $1.660 \pm 0.180$ \\
\bottomrule
\end{tabular}
\end{footnotesize}

%%%%%%%%%%%%%% Table 2 %%%%%%%%%%%%%%%%%

\begin{flushleft}
\textit{(iii) $APE_{90}$}
\end{flushleft}
\vspace*{-.5cm}
\setlength{\tabcolsep}{2pt} 
\begin{footnotesize}
\begin{tabular}{lc|ccc}
{} &              ANN &              BNN &            BNN$_{90\%}$ &BNN$_{80\%}$  \\
\midrule
Pumps [MWh]                 &  $12.450 \pm 0.530$ &  $14.830 \pm 0.510$ &  $12.280 \pm 0.310$ &  \textbf{11.480}$ \pm 0.230$ \\
Heating supply, Other [MWh] &  $20.400 \pm 1.480$ &  $22.300 \pm 0.750$ &  $17.160 \pm 0.580$ &  \textbf{15.240}$ \pm 0.610$ \\
Fans [MWh]                  &  $15.810 \pm 1.540$ &  $18.120 \pm 0.770$ &  $14.950 \pm 0.910$ &  \textbf{13.800}$ \pm 1.050$ \\
Heating supply, Elec. [MWh] &  $13.790 \pm 0.810$ &  $15.130 \pm 0.290$ &  $12.470 \pm 0.490$ &  \textbf{11.670}$ \pm 0.640$ \\
Heating supply, Gas [MWh]   &  $18.320 \pm 0.640$ &  $21.440 \pm 0.620$ &  $16.660 \pm 0.610$ &  \textbf{14.970}$ \pm 0.690$ \\
Cooling supply, Elec. [MWh] &   $6.780 \pm 0.560$ &   $7.490 \pm 0.560$ &   $6.920 \pm 0.460$ &   \textbf{6.540}$ \pm 0.320$ \\
Heating demand [MWh]        &   $7.670 \pm 0.550$ &   $8.040 \pm 0.710$ &   $7.260 \pm 0.740$ &   \textbf{6.940}$ \pm 0.770$ \\
Cooling demand [MWh]        &   $4.620 \pm 0.300$ &   $4.980 \pm 0.090$ &   $4.710 \pm 0.090$ &   \textbf{4.610}$ \pm 0.090$ \\
Interior lights [MWh]       &   \textbf{3.840}$ \pm 0.330$ &   $5.050 \pm 0.160$ &   $4.790 \pm 0.170$ &   $4.560 \pm 0.170$ \\
Interior equipment [MWh]    &   $5.320 \pm 0.960$ &   $5.650 \pm 0.200$ &   $4.780 \pm 0.200$ &   \textbf{4.450}$ \pm 0.240$ \\
Water heating, Gas [MWh]    &   \textbf{1.340}$ \pm 0.100$ &   $2.590 \pm 0.260$ &   $2.350 \pm 0.270$ &   $2.210 \pm 0.250$ \\
PV Generation [MWh]         &   \textbf{2.460}$ \pm 0.320$ &   $6.040 \pm 0.100$ &   $4.120 \pm 0.300$ &   $3.530 \pm 0.350$ \\
\bottomrule
\end{tabular}
\end{footnotesize}

\caption{\textbf{Comparison of Bayesian dropout neural network (BNN) and non-bayesian deterministic neural network (ANN).} The performance of the dropout neural network (BNN) is  provided with and without the application of uncertainty-based thresholding (90\%/80\%).}
\label{tab:comparison}
\end{table}

\end{document}

%% file: 01Introduction.tex
\section{Introduction}
\label{sec:intro}
\begin{figure}
\includegraphics[scale=.6]{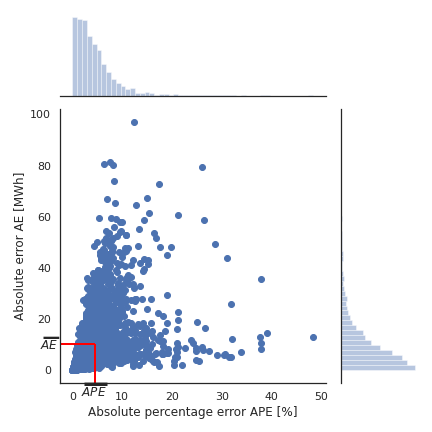}
\caption{\textbf{Distribution of errors of a surrogate model.} The plot shows the error of a surrogate model which emulates the simulation of the heating demand of an office building (see case study in Section \ref{sec:case_study}). While the average absolute error $\overline{AE}$ and absolute percentage error $\overline{APE}$ are low, large errors can occur. This study aims at identifying the large errors using estimates of the surrogate model's uncertainty.}
\label{fig:errors_example}
\end{figure}

A wealth of concepts exist to explore the design of new and existing buildings to improve the building sector's large climate footprint \cite{iea_2019}.   Scaling them is challenging, as usually each building is designed individually corresponding to the cultural context, climatic conditions, surrounding buildings and design preferences. This impedes the distribution of centrally derived design paradigms to the level of individual building projects.\\
Architects and engineers play a vital role to bridge the gap between high-level ideas and individual building projects. Often they use building performance simulation (BPS) to assess the energy and environmental performance of various design options and balance them against design preferences. The computational expense and associated waiting time, however, prohibits an exhaustive design space exploration and optimization. This has led researchers to train machine learning models on simulation input and output data to emulate building simulation models \cite{Westermann2019}. \\
The computational speed of so-called surrogate models has been the basis for a range of innovations in the field of building simulation, for example complex, interactive early design tools (e.g. ELSA \cite{jusselme_2020}, Building Pathfinder \cite{bpf}, \cite{PaulWestermann2020}), faster optimization algorithms \cite{waibel2019building}, and detailed design sensitivity and uncertainty analysis \cite{rivalin2018comparison}\cite{RN363}. A recent survey of building designers confirms that a cohort which received realtime feedback from a surrogate model arrived at higher performing building designs \cite{brown2020}.\\
%Introduce the problem
The growing application of surrogate models draws attention towards the robustness of their performance. Studies have shown satisfactory average accuracy on test data \cite{RN351} which can be slightly influenced by the type and the complexity of inputs \cite{westermann2020weather} and the selection of outputs \cite{PaulWestermann2020}.\\
Nonetheless, average errors computed on test data can be deceiving (see Figure \ref{fig:errors_example}). Test data usually consists of design samples distributed uniformly in the design space and may not reflect the portion of the space the building designer is interested in. Large errors on specific building designs may occur (heteroscedasticity of the errors), affecting important design choices and potentially lowering the energy performance of the final building.\\

\begin{figure}
\includegraphics[scale=.5]{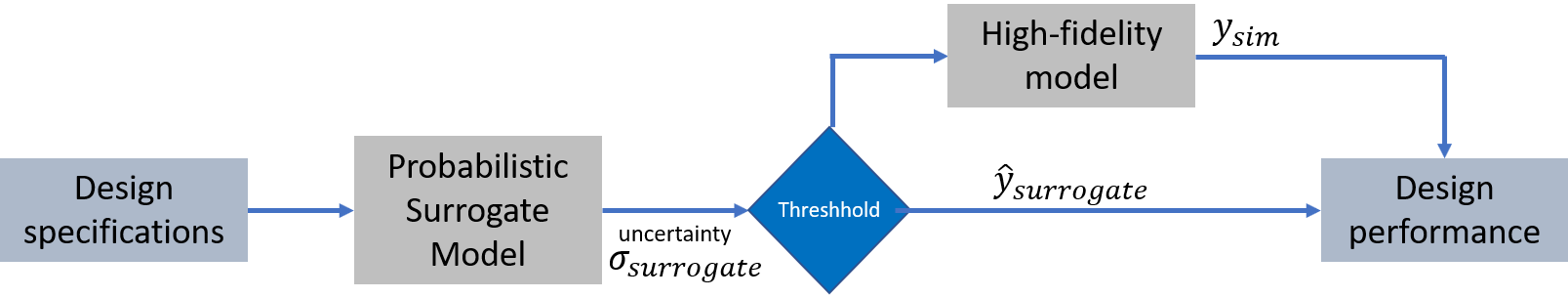}
\caption{\textbf{Uncertainty estimates to link high-fidelity model and a surrogate model.} The surrogate model provides both a performance estimate $\widehat{y}_{surrogate}$ and an uncertainty estimate $\widehat{\sigma}_{surrogate}$. If the uncertainty is large, a high-fidelity model (e.g. a building energy simulation) is querried to produce accurate estimates $y_{sim}$ of an engineering design (e.g. a building). Compare to \cite{filos2019systematic}}
\label{fig:concept}
\end{figure}

Bayesian methods offer a framework to quantify the uncertainty stemming from the inadequacy of an approximate model (epistemic uncertainty) and recent developments in Bayesian deep learning (BDL) managed to integrate them into large machine learning models \cite{kendall2017uncertainties}\cite{damianou2013deep}. As a result BDL models can express for which inputs their estimates are uncertain. In our case, a Bayesian surrogate model produces a building performance estimate as a probability distribution, where the entropy or variance of that distribution allow us to quantify the uncertainty. The architect or building designer is therefore provided with a level of confidence in the performance results and thus, can define uncertainty thresholds above which the high-fidelity model, here the BPS tool, is queried to guarantee high confidence results (see Figure \ref{fig:concept}). 

In this study, we explore two different Bayesian models,  Bayesian neural networks \cite{gal2016dropout} and stochastic variational Gaussian process models \cite{hensman2013gaussian}, to quantify epistemic uncertainty in surrogate models (see Section \ref{sec:Background}). We benchmark the overall accuracy against non-Bayesian surrogate models, validate the quality of the uncertainty estimate, and quantify how a \textit{hybridization} of fast but approximate, and slow but accurate models reduces the error of a surrogate model while computational costs increase only slightly (see Section \ref{sec:results} ff.).

%% file: 02Background.tex
\section{Background}
\label{sec:Background}

\subsection{Motivation for surrogate modelling}
The fundamental motivation to emulate a physics-based high-fidelity model is computational efficiency; simulation outputs can be estimated many orders of magnitude faster, effectively in real-time. This allows a holistic design space analysis which would be infeasible with a slow simulation model. Various applications are found in the building domain as well as other domains \cite{reichstein2019deep}\cite{wang2007review}:
\begin{itemize}
\item General design space exploration: The relationship between design parameters and performance is interactively explored to improve the user's understanding of the design problem \cite{RN401}\cite{brown2020}. This can happen on the single building level or on the urban level \cite{vazquez2019deep}. Often a parallel-coordinates plot is used to visualize the multi-dimensional problem space \cite{PaulWestermann2020}.
\item Design optimization: The surrogate model is trained and queried to accelerate iterative optimization algorithms \cite{RN397}\cite{RN389}\cite{bre2020efficient}. Adaptively training the surrogate model on new simulation samples collected at each optimization iteration can further increase optimization performance \cite{waibel2019building}. 
\item Sensitivity analysis: The surrogate model is used to run the extensive sampling (thousands of simulation runs) required for global sensitivity analysis methods \cite{rivalin2018comparison}.
\item Design uncertainty analysis: Several types of uncertainties exist during the building design process - caused by undetermined design parameters, uncertain contextual parameters (e.g. surrounding buildings, carbon factors, etc.), and vague design constraints \cite{RN414}. This uncertainty is often quantified using Monte Carlo sampling methods, where samples from uncertain parameter distributions are drawn and simulated to quantify how that parameter uncertainty propagates to building performance uncertainty. With a surrogate model, these uncertainties can rapidly be calculated and updated throughout the design process \cite{RN363}.
\item Simulation model calibration: An accurate calibration of a simulation model is required to assess retrofit design choices for an existing building. The calibration, i.e. the process of determining uncertain building parameters, often relies either on iterative optimiziation algorithms \cite{coakley2014review}, or on Bayesian calibration of these uncertain parameters \cite{RN383}. In both cases simulations are iteratively run to closely match simulation outputs with measured sensor data by adjusting the unknown parameters. One can use surrogate models to reduce the computational limitations of these approaches. Note that simulation model calibration can be done both for a specific building \cite{heo2012calibration} or for multiple buildings \cite{sokol2017validation}. The latter commonly requires an archetype model whose parameters are repeatedly calibrated using measurements of the considered buildings \cite{kristensen2018hierarchical}.
\end{itemize}

\subsection{Surrogate model derivation}
In surrogate modelling, we fit a machine learning model to a simulation dataset $D = \lbrace x_n, y_n \rbrace_{n=1}^N =(X,Y)$, where the inputs $X$ correspond to the simulation parameters and $Y$ to real-valued outputs of the simulation run \cite{wang2007review}.\footnote{Also categorical outputs can be considered but practical examples are lacking in building simulation literature.} In the case of building energy surrogate models, the simulation parameters are the building design parameters (e.g. insulation value of the walls) and the outputs are the simulated building performance metrics like total energy consumption or greenhouse gas emissions \cite{Westermann2019}. Studies also exist with time series outputs, like hourly energy demand \cite{vazquez2019deep}.\\
For deriving the surrogate model the modeller first needs to carefully specify the design problem, which includes choosing the free design parameters and the performance objectives as well as all other important contextual parameters (surrounding buildings, etc.). Then simulations are run to create the simulation dataset $D$. The idea is to gain maximum information about the design space (the collection of all possible parameter combinations) per simulation run. Tailored sampling schemes exist, called design-of-experiment methods \cite{garud2017design}, e.g. Latin-Hypercube-sampling that uniformly distributes samples in the multidimensional input space. The number of samples must be specified (e.g. 10-1000 samples per parameter dimension \cite{Westermann2019}) and is adjusted if model accuracy on test samples is too low.\\
Metrics like the coefficient of determination ($R^2$), the mean absolute percentage error ($MAPE$), or the root-mean-squared-error ($RMSE$) can be used. Based on \cite{RN351} and \cite{PaulWestermann2020}, accuracies of $R^2>0.99$ are feasible when estimating annually aggregated performance metrics, e.g. heating demand, but they can be significantly lower when more complex performance metrics are estimated. \\
As mentioned above, surrogate model accuracy is commonly reported as one metric, implying homoscedastic errors. This may not always hold, i.e. the errors may depend on the choice of inputs (heteroscedasticity). By using Bayesian deep learning \cite{kendall2017uncertainties}, we aim to train surrogates that are aware of where in the design space, i.e. for which kind of building designs $x\in X$, the model is uncertain and may produce large errors.

\subsection{Uncertainty in surrogate models}
 The true simulation function $y=f(x)$ is not explicitly available. We use the surrogate model to find an estimate $\hat{f}$ to approximate that function. The central root of uncertainty in surrogate modelling is how plausible the determined $\hat{f}$ is (model uncertainty or \textit{epistemic} uncertainty) \cite{kendall2017uncertainties}. For the most part, this uncertainty is caused by the training set $D = (X,Y)$ which contains only a finite set of points within the space of possible simulation parameter combinations $X$ (the design space) and associated building performance $Y$. Theoretically, epistemic uncertainty can be reduced to zero given more and more data \cite{kendall2017uncertainties}.\\
We consider the problem of surrogate modelling as free of \textit{aleatoric uncertainty}, which represents the noise inherent in observations.\footnote{In the case of sensor data, this can correspond to sensor noise. Here, we consider simulation runs to be deterministic, i.e. the impact of numerical noise to be small. In the case of numerical building simulation, here EnergyPlus \cite{crawley2001energyplus}, this corresponds to the numerical noise of solving the thermodynamic-based differential equations.} Therefore, we only deal with epistemic uncertainty. We propose that quantifying this uncertainty can be a powerful aid in surrogate modelling as it acknowledges that we have to train our model with a limited number of simulation samples that represent a fraction of the design space, which makes the surrogate model uncertain. Bayesian modelling now allows us to reason under that uncertainty, while still benefiting from the advantages of surrogate modelling, i.e. the computational efficiency for large scale design space exploration.

%Machine learning models are usually developed from data as deterministic machines that map input to output using a point estimate of parameter weights calculated by maximum-likelihood methods. However, there is a lot of statistical fluke going on in the background. For instance, a dataset itself is a finite random set of points of arbitrary size from a unknown distribution superimposed by additive noise, and for such a particular collection of points, different models (i.e. different parameter combinations) might be reasonable. Hence, there is some uncertainty about the parameters and predictions being made. Bayesian statistics provides a framework to deal with the so-called aleoteric and epistemic uncertainty.

%In this paper we use Gaussian Processes and Bayesian Neural networks as Bayesian machine learning methods. The latter should not be confused with Bayesian networks, a causal network of probabilities...

%% file: 03Methodology.tex
\section{Bayesian modelling for surrogates}
\begin{figure}
\includegraphics[scale=.5]{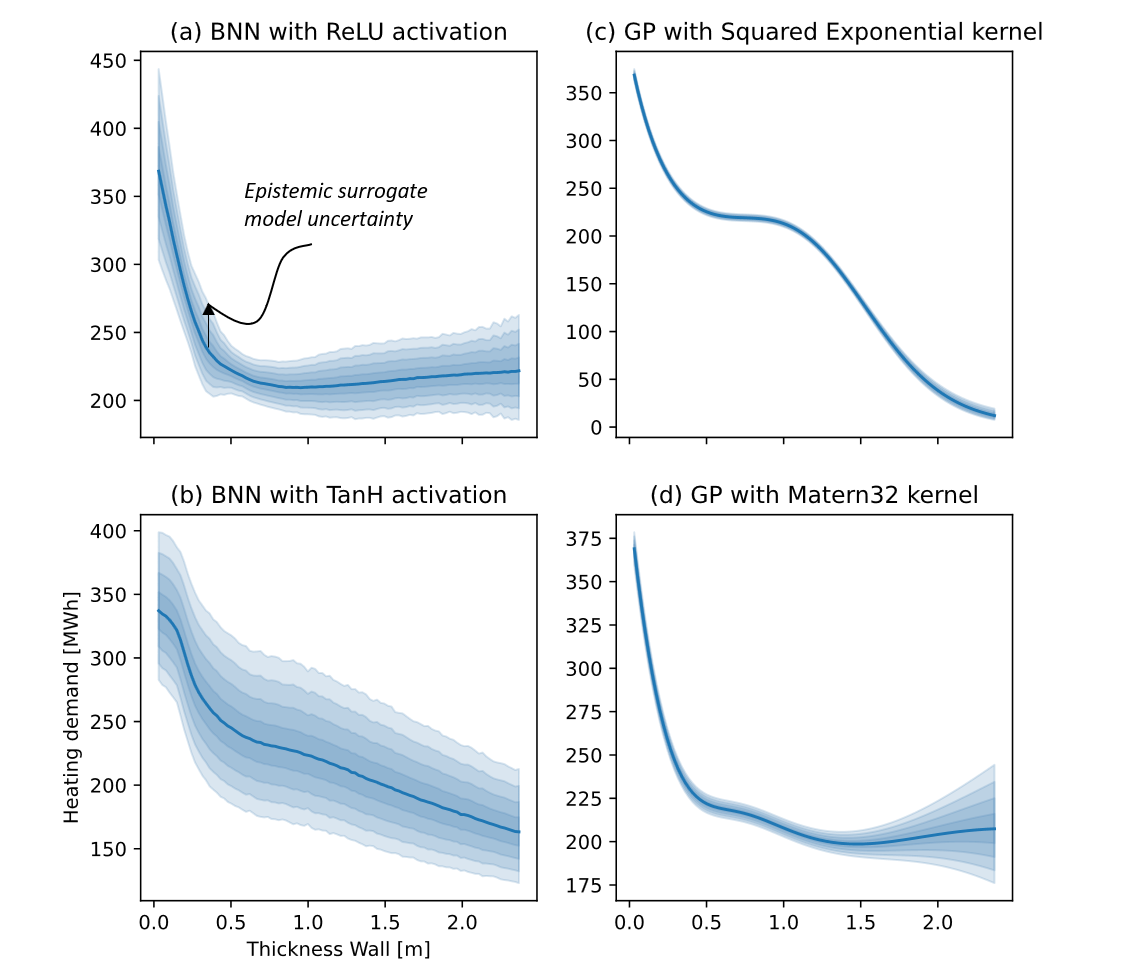}
\caption{Bayesian neural network heating demand estimate and associated epistemic uncertainty. In particular, out-of-sample the uncertainty of the surrogate model is large. Out-of-sample is that part of the design space, where no (or few) simulations to train the surrogate model on a were collected.}
\label{fig:example_functionfamily}
\end{figure}

Bayesian probability theory offers us grounded tools to quantify model uncertainty \cite{rasmussen2004gaussian}.\\
To understand the core idea of Bayesian modelling, we consider a parametric model $y=f(x,\Theta)$, where $x$ is the input, $f$ is a space of possible models (see Figure \ref{fig:example_functionfamily}) and $\Theta$ is the set of model parameters (for example the weights in a neural network). Instead of finding a single $\Theta$, in Bayesian modelling we search for a collection of $\Theta$, which likely has produced the output $Y$ given $X$. In our case we search for a collection of surrogate models with different weights.\\
The Bayesian theorem, as shown in Eq. \ref{eq:bayesian_law}, is applied to find a collection which likely has produce $Y$ given $X$. Based on our prior knowledge on the distribution of the model weights $p(\Theta)$ and combined with the likelihood function $p(Y|X,\Theta)=\prod_{n=1}^{N}p(y_n|x_n,\Theta)$, which quantifies the probability that a specific model parameter set generated the observations $(X,Y)$, the posterior of the model parameters can be computed.
\begin{equation}
\label{eq:bayesian_law}
p(\Theta|Y,X)=\frac{p(Y|X,\Theta)p(\Theta)}{p(Y|X)}
\end{equation}
where $p(Y|X)$ is called the marginal likelihood. It represents the probability of the observed data given the model $f$ with all possible model parameters. It is a scalar that normalizes the posterior. Given the posterior, we can now infer about future data in form of a predictive distribution:
\begin{equation}
p(y_*|x_*,X,Y)=\int p(y_*|x_*,\Theta)p(\Theta|X,Y)d\Theta
\end{equation}
The mean and variance or entropy can be derived, where the latter two provide information on the uncertainty in the estimated values. In the building surrogate modelling setting, we predict an expected building performance, e.g. annual heating demand, and an associated uncertainty given building design parameters, e.g. the thickness of the wall (see Figure \ref{fig:example_functionfamily}).

%Came at high computational cost

\subsection{Variational inference}
\label{sec:VI}
The true posterior of the weights $p(\Theta|Y,X)$ however, is commonly intractable. This is particularly the case in the big data regime when more complex models are required \cite{hensman2013gaussian}. In the small data regime (below a few thousand samples) posterior inference with a standard Gaussian Process Bayesian model is feasible and was successfully applied for building surrogate models \cite{ostergaard2016building}\cite{heo2012calibration}. However, with increasing complexity, for example more inputs and outputs (e.g. \cite{westermann2020weather}), standard GPs have major shortcomings:
\begin{itemize}
\item The model complexity is limited as it only consists of one layer, i.e. the outputs of the GP are not used as inputs to another GP. This prohibits modeling hierarchical structures and abstract information \cite{damianou2013deep}.
\item Computational cost increase with the cubically ( $\mathcal{O}(n^3)$) with the number of samples $n$. This prohibits increasing the size of the surrogate model training set to improve the model accuracy (for example to train a complex, tailored kernel with many hyperparameters \cite{rasmussen2004gaussian}).
\end{itemize}

Instead, recent advances in variational inference (VI) allow us to approximate the true posterior of $\Theta$ in big data problems \cite{blei2017variational}. We pick an approximate variational distribution over the (latent) model parameters $q_{\nu}(\Theta)$ with its own variational parameters $\nu$. Now we search for $\nu$ that minimizes the divergence to the true posterior which is quantified by the so-called \textit{Kullback-Leibler (KL) divergence}. Thereby the marginalization, i.e. the integration required to calculate the true posterior, is turned into an optimization problem which is often easier to solve. The approximative distribution of $q$ can be used to form predictions about unseen samples.\\

Scalable variational inference methods have been developed both to do approximative inference with Bayesian neural networks (BNN) \cite{kendall2017uncertainties} and with sparse variational Gaussian process (SVGP) models \cite{hensman2013gaussian}. The two approaches are introduced in the following section. 

\begin{figure}
\includegraphics[scale=.5]{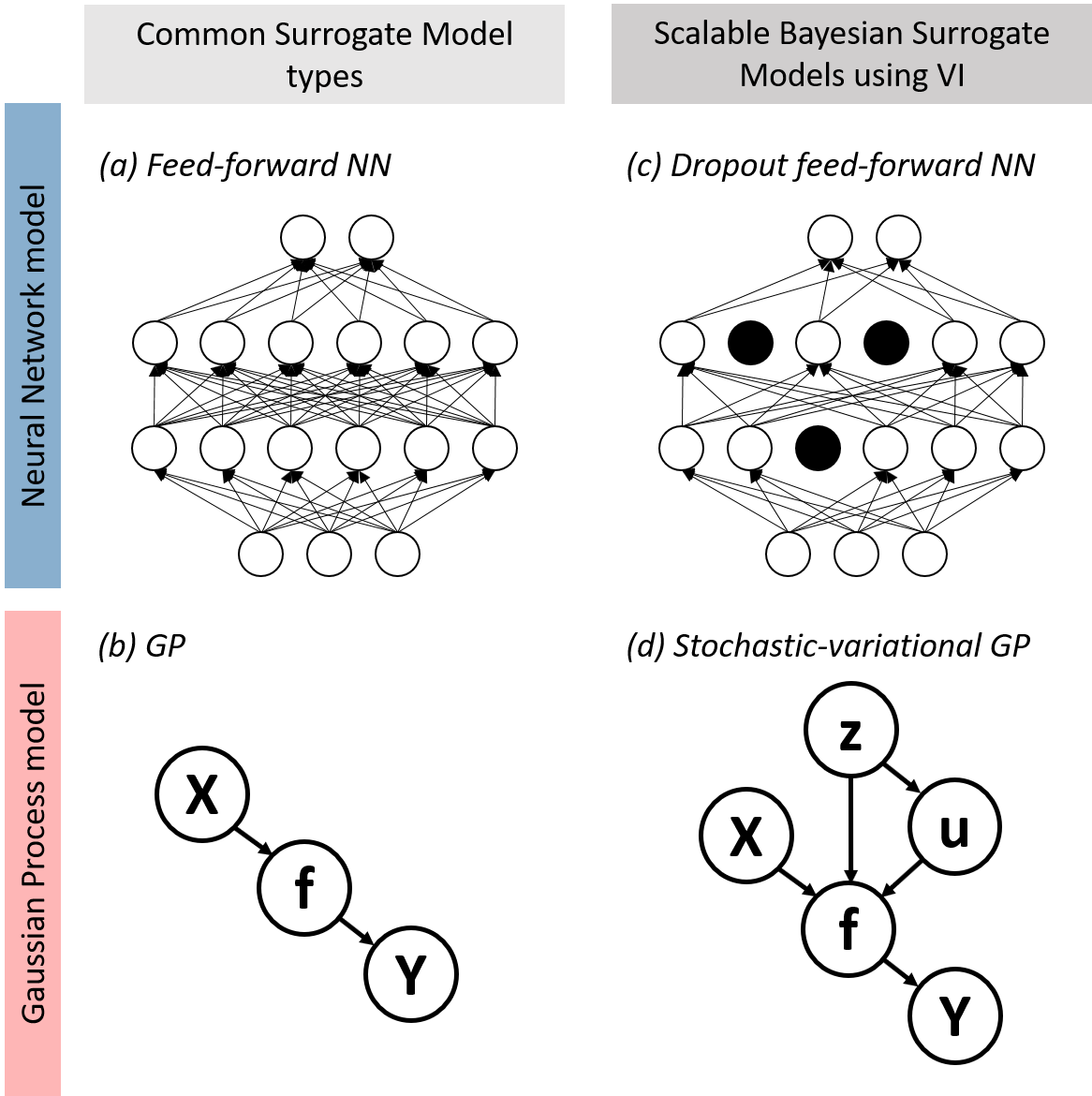}
\caption{\textbf{Considered variational-inference approaches to turn existing surrogate modelling architectures into scalable Bayesian models \cite{hensman2013gaussian}\cite{gal2016dropout}.}}
\label{fig:new_surr}
\end{figure}

\subsection{Deep Bayesian Neural Networks}
The concept of a Bayesian neural network (BNN) is an extension of standard network architectures (e.g. feed-forward neural network, convolutional neural network, or recurrent neural network) to follow the Bayesian modelling paradigm \cite{neal1995bayesian}. In a BNN we sample the neural network weights from a prior distribution rather than having a single fixed value as in normal neural networks, for example from a Gaussian ${\Theta} \sim N (0, I)$ \cite{gal2016uncertainty}. Instead of optimising the network weights directly we average over all possible weights, called marginalisation. Given the stochastic output of the BNN $f^{\Theta} (x)$, we receive a model likelihood $p(y|f^{\Theta} (x))$. Based on the dataset $D$, Bayesian inference is used to compute the posterior over the weights $p({\Theta}|X, Y)$. This posterior captures the set of all plausible model parameters. This  distribution  allows predictions on  unseen  data. %For regression tasks we often define our likelihood as a Gaussian with mean given by the model output: $p(\widehat{y}|f^{\Theta} (\widehat{x})) = N (f^{\Theta} (\widehat{x}), \sigma^2 )$, with an observation noise scalar $\sigma$. %redo that sentence!!

As mentioned above the exact posterior is intractable, and different approximations exist \cite{gal2016dropout}\cite{pearce2018uncertainty}. In these approximate inference techniques, the posterior $p({\Theta}|X, Y)$ is fitted with a simple distribution $q(\Theta)$. Here we consider the Dropout variational inference approach as it has shown great performance when benchmarked against other methods \cite{gal2016dropout}\cite{filos2019systematic}.

\subsubsection{Dropout variational inference}
Dropout variational inference is a variational inference approach, i.e. it allows to find a $q^{\ast}_{\nu}(\Theta)$ that minimises the Kullback-Leibler divergence to the true model posterior, that neither requires to change the architecture of common network architectures nor to change the optimisation algorithm for training the network \cite{gal2016uncertainty}. The inference of the posterior is done by training a model which uses stochastic dropout on every neuron layer \cite{srivastava2014dropout} (see Figure \ref{fig:new_surr}). This stochastic dropout is also used to remove neurons when performing predictions. By repeating the predictions (stochastic forward passes), we create a distribution of outputs, which was shown to minimize the KL divergence \cite{gal2016uncertainty}.\\
This KL divergence objective is formally given in the following, where we approximate $p({\Theta}|X, Y)$ with $q(\Theta)$  \cite{gal2016uncertainty}\cite{kendall2017uncertainties}:
\begin{equation}
\L(\Theta, p) = \text{-}\frac{1}{N} \sum_{i=1}^{N} \text{log} \hspace*{.1cm} p(y_i |f^{\widehat{\Theta_i}}(x_i)) + \frac{1\text{-}p}{2N}||\theta||2
\end{equation}

with $N$ data points, dropout probability $p$, weight samples $\widehat{\Theta_i} \sim q^{\ast}_{\nu}(\Theta)$,
and $\theta$ the set of the sample distribution's parameters to be optimised (weight matrices in the dropout case). Note that for each data point in the training set dropout is applied, which provides us with $N$ samples of $\Theta_i$.\\
When performing dropout variational inference the $T$ stochastic forward passes provide us with the epistemic uncertainty given by the variance $Var(y)$:
\begin{equation}
Var(y) \approx \frac{1}{T}\sum_{t=1}^{T}f^{\widehat{\Theta_t}}(x)^T f^{\widehat{\Theta_t}}(x_t) \text{-} E(y)^T E(y)
\end{equation}
with predictions in this epistemic model done by approximating the predictive mean: $E(y)\approx \frac{1}{T}\sum_{t=1}^{T} f^{\widehat{\Theta_t}}(x)$. Note that in this formulation we assumed no noise inherent in the data and therefore, $Var(y)$ is zero when we have no parameter uncertainty.

\subsubsection{Model architecture and implementation}
\label{sec:BNN_archi}
We implemented a dropout neural network using the Keras Tensorflow API \cite{chollet2015keras}\cite{abadi2016tensorflow} based on the work from Gal and Gahramani \cite{gal2016dropout}. Our network is a feed-forward neural network with 2 hidden layers of 512 neurons which are activated with a leaky rectified linear (ReLU) function. Training was done within 1200 epochs using a batch size of 128 samples. A dropout rate of 5\% was set. All mentioned parameters ($n_{layers}\in [1,2,3]$, $n_{neurons}=[256, 512, 1024]$, dropout rate $\in [5\%, 10\%, 20\%]$) were analysed in a 5-fold cross-validation. The model with the highest accuracy on the test set was picked. Furthermore, we analysed the impact of the dropout rate on the uncertainty quality (see Section \ref{sec:evaluation_criteria}), but no significant change in the performance was observed, which agrees with the observation from \cite{gal2016dropout}, that the uncertainties of models with different dropout rate converge with the training progress.

\subsection{Gaussian Processes in the Big Data regime}
Gaussian Processes models are attractive for non-parametric Bayesian modelling \cite{rasmussen2004gaussian}. They use a Gaussian Process prior for a stochastic, latent function $f$ to describe the relationship between $X$ and $Y$ (see Figure \ref{fig:new_surr}). The function values $f(x)$ are assumed to be sampled from that Gaussian with zero mean and covariance matrix $K$, i.e. $f \sim \mathcal{N}(0, K)$. The choice of covariance function impacts various aspects of the GP model and also determines which model parameters $\Theta$ to be tuned. These model parameters are optimized when training the GP model.\\
However, given the above-mentioned limitations of standard Gaussian Process models (see Section \ref{sec:VI}), sparse GP approximations have been developed to handle large datasets by lowering the computational complexity to $\mathcal{O}(nm^2)$ \cite{titsias2009variational}\cite{bauer2016understanding}.\footnote{This blog post provides a summary on the history on sparse Gaussian Process models: \url{https://www.prowler.io/blog/sparse-gps-approximate-the-posterior-not-the-model}.} They rely on the use of inducing variables (or pseudo-inputs), i.e. a reduced set of latent variables with size $m<<n$ to represent the actual data set $D$ with $n$ samples. The $m$ inducing points are GP realisations $u=f(z)$ at the inducing locations $Z$ which are in the same space as the observed inputs $X$ (but not necessarily part of $X$). When training the SVGP, the locations of the inducing points $Z$ and the covariance parameters $\Theta$ are optimally chosen to minimize the KL divergence. Important is that the locations $Z$ are parameters to shape the variational approximate distribution $q(f)$, rather than being part of the model parameters $\Theta$, i.e. the covariance function with parameters $\Theta$ are calculated \textit{for} the inducing locations $Z$.\\

In comparison to sparse GPs \cite{titsias2009variational}, stochastic variational GPs \cite{hensman2013gaussian} allow mini-batch training which further reduces computational complexity to $\mathcal{O}(n_{batch}m^2)$. Since \cite{hensman2013gaussian} and others, deep Gaussian Process models have been developed, too, but are not considered in this study as our case study data set is still of limited size and complexity \cite{damianou2013deep}\cite{salimbeni2017doubly}. However, our SVGP model may be regarded as a one-layered deep GP \cite{svendsen2020deep}.

\subsubsection{Model architecture and implementation}
\label{sec:gp_archi}
Here we train a one-layered stochastic variational Gaussian Process model on batches with 100 samples with a Matern32 kernel covariance function using the GPy implementation based on \cite{hensman2013gaussian}\cite{gpy2014}. Again we ran a 5-fold cross validation to pick the covariance function as also a simpler squared-exponential kernel was analysed. Furthermore, although the observed dataset is deterministic, we considered a fixed noise level in the model ($\approx 0.001\%$ of the mean absolute value of the outputs) as it produced much more accurate models. This implies that a deep Gaussian process may be a better choice than o one-layered SVGP.

\section{Case Study: Surrogate models for the design of net-zero energy buildings}
\label{sec:case_study}
\subsection{Objective}
We use a case study on a popular topic in the building domain, the design of buildings with net-zero energy demand, to train and assess the two Bayesian model types introduced above. It shall serve as an example showcasing the use of both model types for building surrogate modelling, but should not be considered as an exhaustive comparison of the two. For that purpose the reader is referred to other studies instead, e.g. \cite{filos2019systematic} or \cite{salimbeni2017doubly}.

\subsection{Case study building}
We emulate simulation outcomes of one archetype building contained in the NetZero navigator project \cite{PaulWestermann2020}. The NetZero navigator projects hosts building simulation surrogate models on a web-platform, which enable to predict building energy consumption of archetype buildings given a large set of building design parameters in real time. So far the platform relied on common deterministic neural network surrogates, whose building performance estimation accuracy was validated on separate building designs not contained in the training data. All the  simulation runs for training and testing were collected with the well-known building performance assessment program EnergyPlus \cite{crawley2000energyplus}. To date, no design-specific uncertainty estimate is produced to tell the user when the surrogate model estimate is not trustworthy.\\ 

For this case study, we look at a medium office archetype building, where 35 design parameters are free to choose and the building energy performance is quantified by 12 separate performance metrics. The office architecture is based on work from the US DOE Canmet-Energy which derived commercial prototype building models. The development of the parameter set, the choice of performance metrics, and software to generate the (parametric) simulation data set, however, was developed individually for that project, where the parameter ranges are directly based on requirements in the Canadian building sector \cite{NECB}. The mechanical systems are parametrized to capture a wide variety of configurations allowing direct manipulation of the air-side system (incl. heat recovery ventilation, various pump efficiencies) and plant equipment performance of various systems (heat pump, electric resistance heater, biogas furnace, natural gas furnace, air conditioning system). This allows us to explore a large HVAC system design space on a high-level (incl. multi-system setups). All details on the building may be found in \cite{PaulWestermann2020}.

\subsubsection{Data set and transformations}
We sample the large design space using 10'000 simulation runs, where each individual parameter combination was picked using the space-filling Latin-Hypercube-sampling (LHS) \cite{garud2017design}. Similarly, we run additional 3000 simulations and use it as a separate test set. Each individual building simulation run took approximately 2 minutes and 10 seconds using 1 CPU and 4 GB RAM, but varied depending on the parameter choices.\\
Prior to training, we standardized the uniformly distributed inputs with different ranges to be normally distributed with zero mean. Furthermore, we transformed the 12 output variables to also be close to a normal distribution. Therefore, adaptive Box-Cox transformations was applied \cite{box1964analysis}. It adaptively finds transformation parameters to transform various kinds of distributions (here of 12 different outputs) to normal distributions. This, in particular, increased the accuracy of the multi-output neural network compared to other transformations.

\subsection{Evaluation criteria}
\label{sec:evaluation_criteria}
We evaluate the models with regard to multiple objectives: (i) the model accuracy, (ii) uncertainty accuracy , (iii) the effectiveness of uncertainty-estimate-based issue-raising.
\subsubsection{$R^2$ score, MAPE and RMSPE score to quantify overall surrogate accuracy}
Our error metrics cover common metrics in the field, i.e. the R$^2$ \cite{RN351} and the Mean Absolute Percentage Error (MAPE) \cite{RN358}. Furthermore, we added the $APE_{90}$ error, i.e. the 90st-percentile of the absolute errors sorted by ascending magnitude, to quantify the robustness of the surrogate model \cite{RN399}.

\begin{equation}
\text{R}^2(Y, \hat{Y}) = 1 \text{-} \frac{\sum_{i=1}^{n} (y_i \text{-} \hat{y}_i)^2}{\sum_{i=1}^{n} (y_i \text{-} \bar{Y})^2}
\end{equation}
%\begin{equation}
%\text{nMBE}(y, \hat{y}) = \frac{1}{n}  \frac{\sum_{i=1}^{n}| y_i \text{-} \hat{y}_i |}{\bar{y_i}}, \text{and}
%\end{equation}
\begin{equation}
\text{MAPE}(Y, \hat{Y}) = \frac{1}{n} \sum_{i=1}^{n} \frac{| y_i \text{-} \hat{y}_i |}{y_i}
\end{equation}
%\begin{equation}
%\text{RMSPE}(y, \hat{y}) = \sqrt{\frac{1}{n} \sum_{i=1}^{n} \left(\frac{y_i \text{-} \hat{y}_i}{y_i}\right)^2},
%\end{equation}

where $\hat{Y}$ corresponds to the matrix of predicted values, $Y$ is the matrix of simulated building performance values. When the error term, $Y \text{-} \hat{Y}$ approaches zero, R$^2$ approaches one, and MAPE goes to zero.\\
\subsubsection{Accuracy of the uncertainty estimate}
In a well-calibrated Bayesian model the uncertainty estimates capture the true data distribution, for example a 95\% posterior confidence interval also contains the true simulation outcome in 95\% of the times \cite{kuleshov2018accurate}. Quantifying the level of calibration is a well-known concept in classification \cite{platt1999probabilistic} but has also been used for regression problems recently \cite{scalia2020evaluating}\cite{kuleshov2018accurate}.\\
Formally, we say that the uncertainty estimates of the surrogate model are well-calibrated if
\begin{equation}
\frac{\sum_{n=1}^{N}\lbrace y_t\leq F_t^{\text{-}1}(p) \rbrace}{N} \rightarrow p \text{ for all p}\in [0,1]
\end{equation}
where $F_t$ is the cumulated density function targeting $y_t$ and $F_t^{\text{-}1} = inf\lbrace y:p\leq F_t(y_t) \rbrace$ is the quantile function. Here we consider each prediction as a standard, symmetric Gaussian distribution $\mathcal{N}(\mu(X), \sigma(X))$. \footnote{This is not necessarily true and possibly a recalibration step is required \cite{kuleshov2018accurate}.} The confidence intervals can be computed using the inverse cumulated density function. To assess the calibration quality, we count the fraction of observations in the test data falling in the prediction confidence intervals derived from the quantile function (see Figure \ref{fig:cali_plot}, left).\\

We show the level of calibration of the Bayesian models in Figure \ref{fig:cali_plot} (left), where perfectly calibrated uncertainty estimates would be aligned with the diagonal. To quantitatively compare different calibration curves, one can also compute the absolute difference between the confidence curve and the diagonal, called the calibration error or the area under the curve (AUC) \cite{scalia2020evaluating}. The problem of assessing the calibration quality based on the calibration plot is that it can suggest perfect qaulity with homoscedastic uncertainty estimates, i.e. constant uncertainty estimates for any input. Therefore, we also quantify the \textit{sharpness} of the uncertainty estimates by calculating the overall variance in the uncertainty \cite{kuleshov2018accurate} (see Section \ref{sec:results}).\\

\subsubsection{Discard-ranking to quantify the effectiveness of uncertainty estimates for surrogate model application}
While having accurate uncertainty estimates is the one thing, in building surrogate modelling we are mostly concerned in warning model users, when the model is uncertain and recommend to rather run a simulation instead (see Figure \ref{fig:concept}). Therefore, we derive a ranking of the samples in the test set based on the magnitude in their uncertainty. This provides two conclusions. First, if it strongly overlaps with the actual surrogate model error the uncertainty estimates are an effective heteroscedastic warning mechanism. Second, we can use the ranking to calculate how much the average error can be reduced when referring a certain percentage of most uncertain samples (here 10\% or 20\%) to the high-fidelity simulation program than processing it with a surrogate model.\\
Both aspects are addressed when plotting the mean error computed on discrete percentiles of the test data, where the test data is sorted by the magnitude of the uncertainty. We can compare that curve to the mean error computed using test data sorted by the magnitude of the computed error (oracle ranking). A large distance between the two curves can tell us that the surrogates uncertainty estimates are not helpful to predict when it is inaccurate. Furthermore, by looking at the slope of the curve, we can see by how much the mean error can be reduced if we discard all samples with uncertainties above a certain threshold. 
 

%% file: 04Results.tex
\section{Results}
\label{sec:results}
In this section, we show the results of the case study where we derived uncertainty-aware surrogate models to replaced building energy simulation models.\\
In the case study, we trained two different Bayesian machine learning models to provide epistemic uncertainty estimates, i.e. a deep Bayesian dropout neural network (here abbreviated by BNN) and a stochastic variational Gaussian Process model (SVGP) approach. We scrutinize the performance of both approaches by comparing their predictive accuracy, by comparing the quality of the uncertainty estimates, and by quantifying how effectively the uncertainty estimates allow us to identify possible surrogate prediction errors. 

\begin{figure}
\hspace*{-2.8cm}
\includegraphics[width=1.4\textwidth]{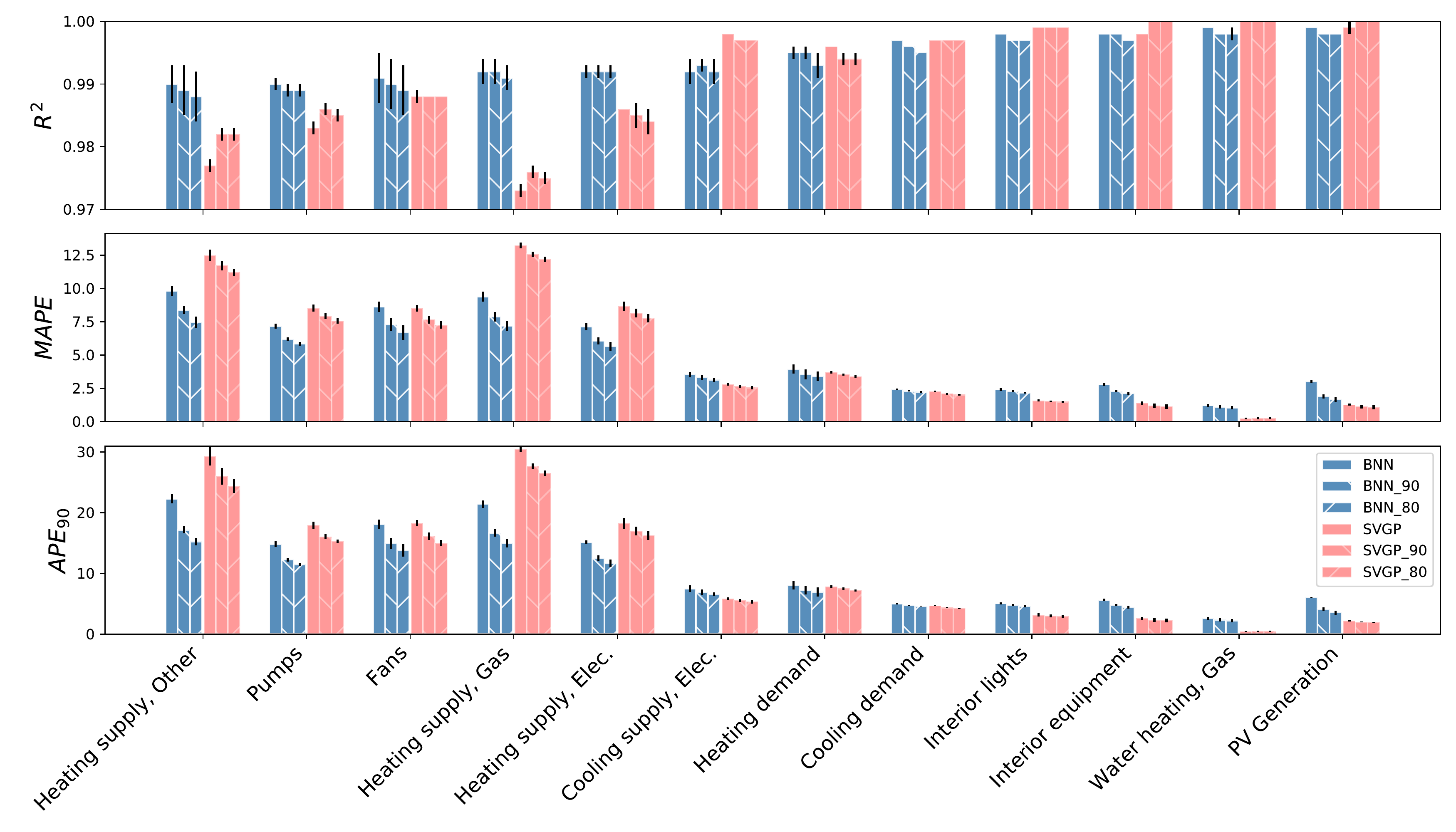}
\caption{\textbf{Summary of results on the use of deep, uncertainty-aware surrogate models.} The plot shows the accuracy, quantified using three different error metrics, of both Bayesian learning approaches for all twelve outputs considered in the case study. The figures also includes performance metrics when we use the uncertainty estimates to identify error prone samples in the test data (textured bars, for details see Section \ref{sec:thresholding}). }
\label{fig:performance}
\end{figure}

\subsection{Model accuracy and uncertainty quality}

\subsubsection{Accuracy}
\label{sec:surrogate_accuracy}
We benchmark the accuracy of the two model types, dropout neural networks and SVGP models. The performance was quantified using three performance metrics as introduced above (see Section \ref{sec:evaluation_criteria}). Each model was trained five times to generate robust results. The results are shown in Figure \ref{fig:performance} and Table \ref{tab:first} in the Appendix; details on the model layout and training process can be found in Sections \ref{sec:BNN_archi} and \ref{sec:gp_archi}).\\
Both considered models reach an accuracy of $R^2>0.97$ on all the outputs, when predicting building performance of buildings contained in the test data. The dropout neural network is more accurate with $R^2>0.99$. Mean percentage errors of $MAPE<13.2\%$ for the GP model and $MAPE<9.82\%$ for the neural network were found. The largest errors occur when estimating the energy demand provided by different heating sources (i.e. the different fuel types), and the air-side system energy demand. Small surrogate model errors are found for the other building performance targets like the photovoltaic (PV) generation, or energy demand for interior lights and equipment.\\
To prove robustness of surrogate model estimates, we specifically look at the largest errors it produces. Therefore, we complement our analysis of the mean absolute percentage error with an analysis of the distribution of the absolute percentage errors observed for each sample in the test data. We extract the 90-th percentile of the distribution as proxy of the largest error found, while ignoring outliers. We abbreviate this metric with $APE_{90}$. $APE_{90}$ errors are found reaching up to $22.3\%$ ($30.5\%$) for the BNN model (GP model), highlighting the demand for increasing the robustness. 

\subsubsection{Uncertainty calibration}
\begin{figure}
\hspace*{-5cm}
\begin{minipage}{.65\textwidth}
\hspace*{2cm}
\includegraphics[scale=.45]{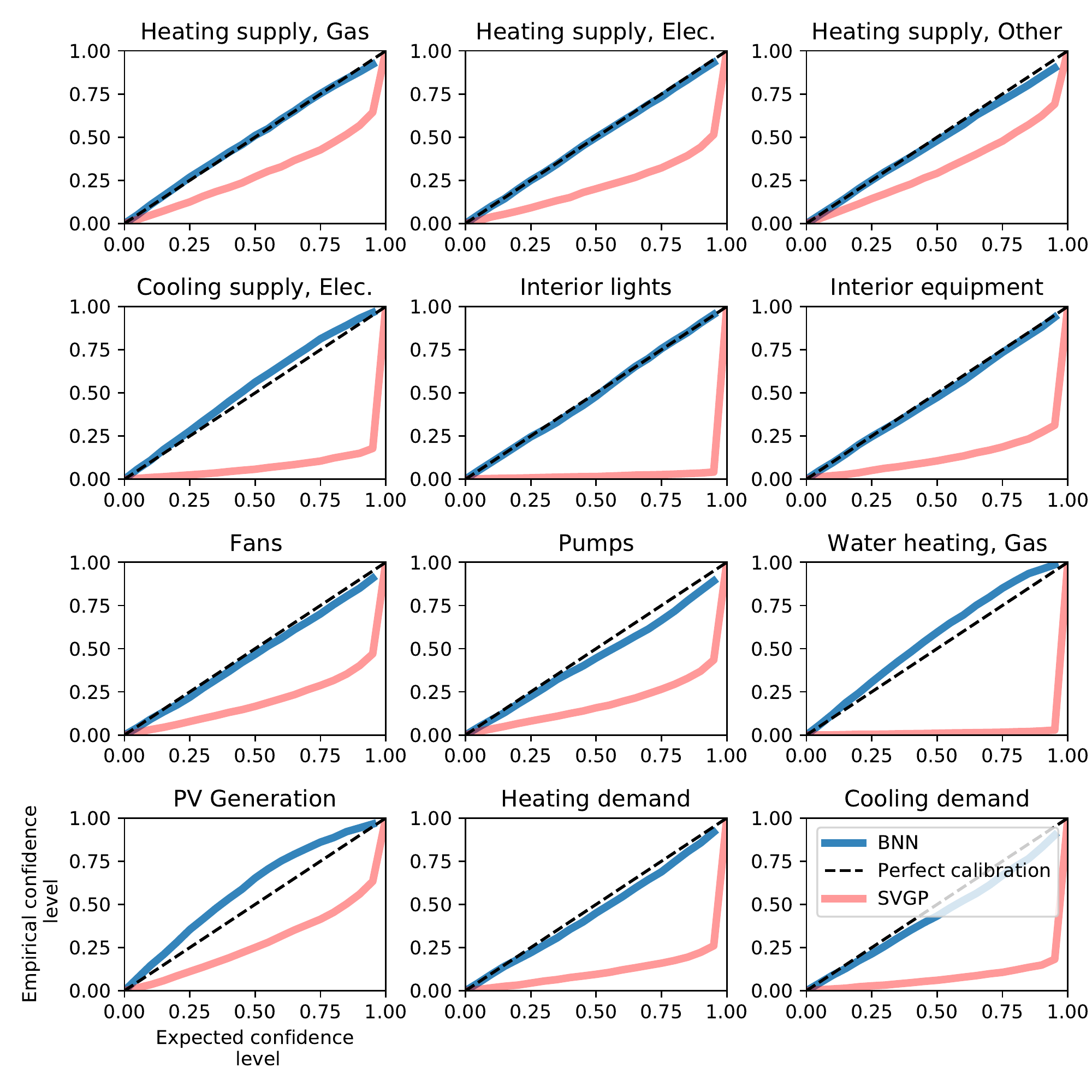}
\end{minipage}
\begin{minipage}{.65\textwidth}
\hspace*{2cm}
\vspace*{.15cm}
\includegraphics[scale=.44]{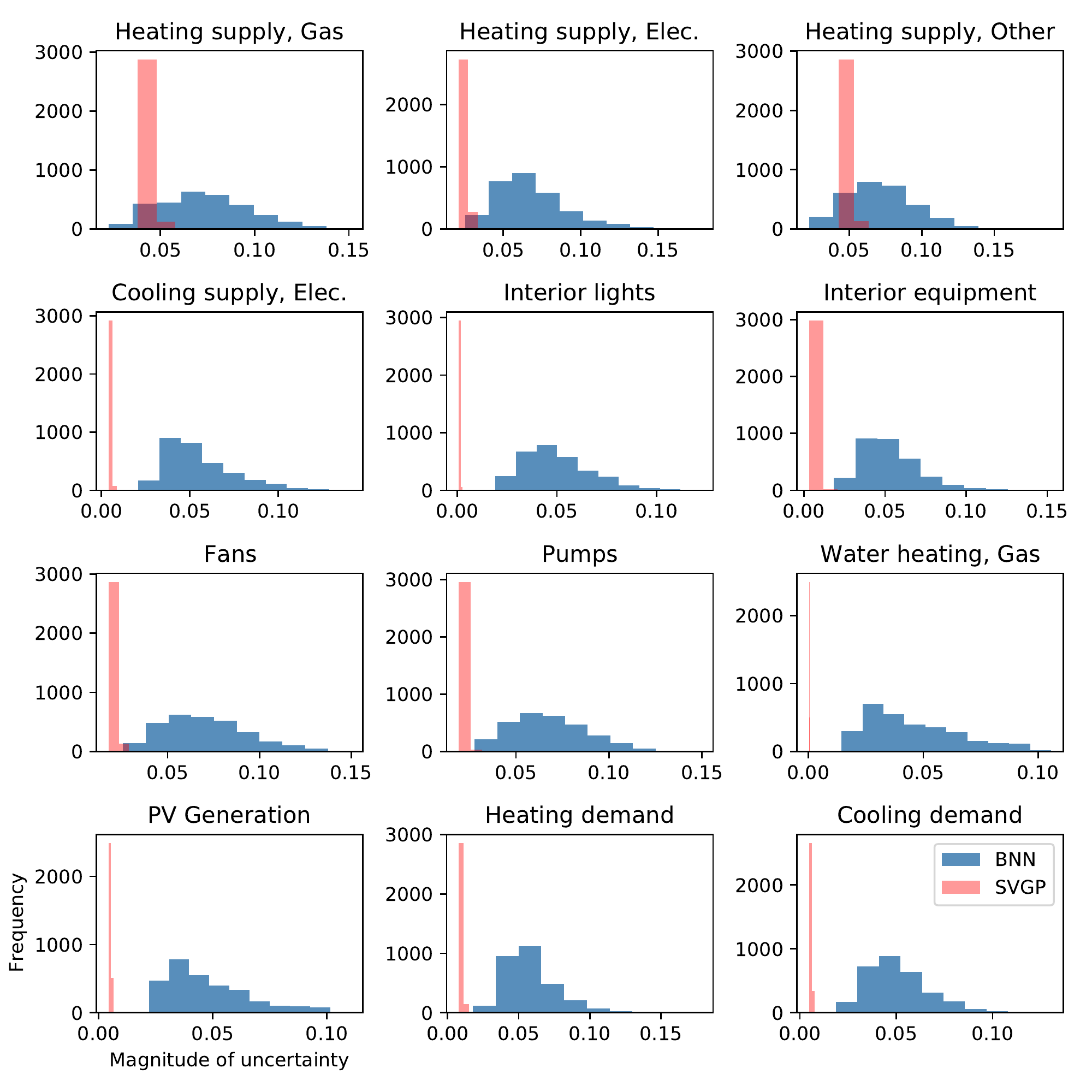}
\end{minipage}

\caption{\textbf{Visualization of the quality of uncertainty estimates of the BNN and the SVGP.} The quality is quantified by how well-calibrated and sharp the uncertainty estimates are. In both regards, the BNN outperforms the SVGP in this study.}
\label{fig:cali_plot}
\end{figure}

When uncertainty estimates are perfectly calibrated, the derived confidence interval, e.g. the 90\% confidence interval, contains the true outcome in the right number of cases, i.e. 90\% of the times for the given example. This is illustrated in Figure \ref{fig:cali_plot}, where we counted for how many times the true simulation outcome was contained in the estimated confidence interval. With a perfectly calibrated Bayesian model the estimated confidence and fraction of the test samples within that interval should perfectly align (dashed line). The region below the dashed line indicates an overly confident model (i.e. confidence bands are too narrow), the region above the dashed line means that the model is too careful having too large confidence bands.\\
We find that the BNN model is well-calibrated, while the GP model is overly confident (Figure \ref{fig:cali_plot}, left). The low quality of uncertainty estimates of the GP model can also be seen on the right, where we display the distribution of all uncertainty estimates collected for predictions of the test data samples. The average magnitude of uncertainty in the GP model indicates its too high confidence, and the small width of the distribution indicates that the uncertainty estimates tend to be homoscedastic, i.e. a similar uncertainty is predicted independent of the model inputs. This width of the distribution is also called the \textit{sharpness} of uncertainty estimates (see Section \ref{sec:evaluation_criteria}). In case of the BNN, the sharpness is better and uncertainty estimates depict a significant level of variance.\\
We can conclude that the uncertainty estimates of the BNN are well calibrated and provide heteroscedastic uncertainty estimates.\\

\subsubsection{Using uncertainty estimates to increase robustness}
\label{sec:thresholding}
\begin{figure}

\begin{minipage}{.49\textwidth}
\hspace*{-3.5cm}
\includegraphics[height=1\textwidth]{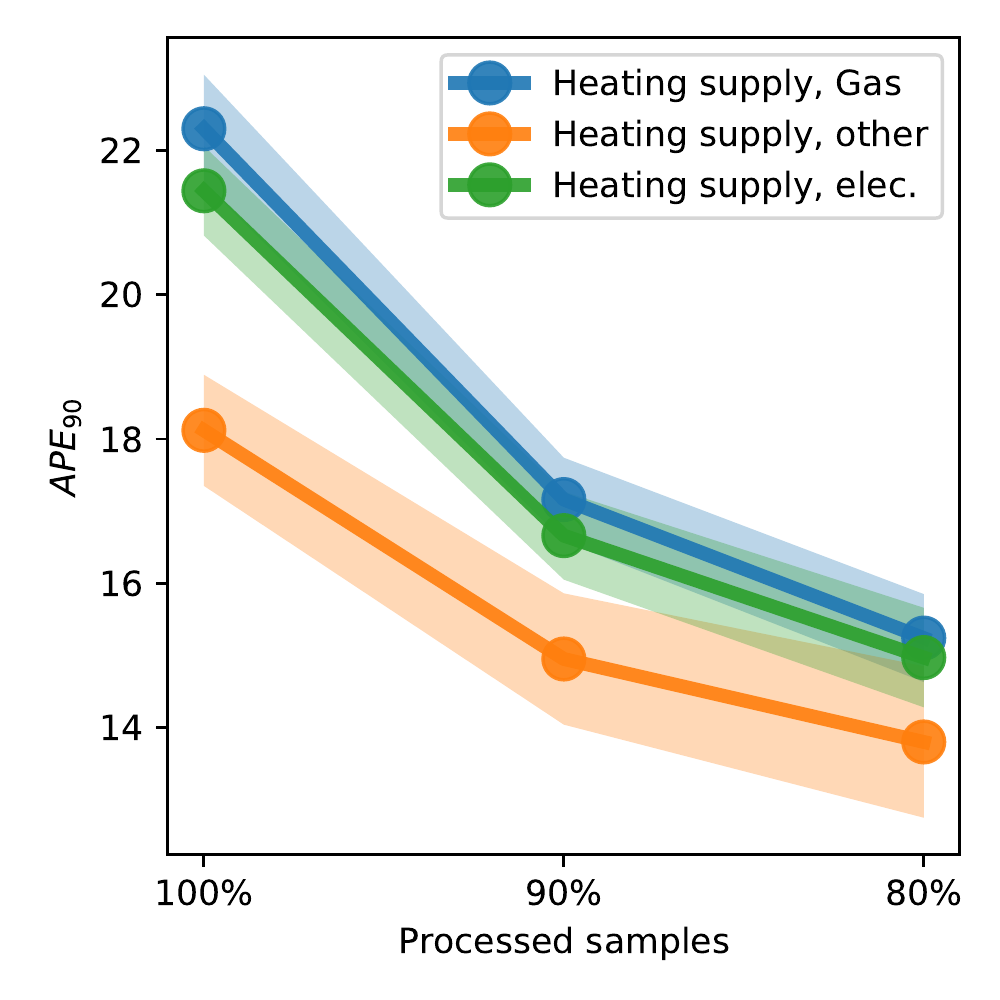}
\end{minipage}
\hspace*{-3.5cm}
\begin{minipage}{.49\textwidth}
\setlength{\tabcolsep}{2pt}
\renewcommand{\arraystretch}{.5}
\begin{footnotesize}
\begin{tabular}{lrr|rr}
\vspace*{.1cm}
Share of samples processed &  90\% &  80\% &  90\% &  80\%  \\
{} &  $\Delta MAPE$ &  $\Delta MAPE$ &  $\Delta APE_{90}$ &  $ \Delta APE_{90}$ \\
\midrule
Heating supply, Gas [MWh]   &                   -16.17\% &                   -23.51\% &                  -22.29\% &                   -30.18\% \\
Heating supply, Elec. [MWh] &                   -15.10\% &                   -20.70\% &                  -17.58\% &                   -22.87\% \\
Heating supply, Other [MWh] &                   -14.66\% &                   -23.83\% &                  -23.05\% &                   -31.66\% \\
Cooling supply, Elec. [MWh] &                    -6.48\% &                   -11.27\% &                   -7.61\% &                   -12.68\% \\
Interior lights [MWh]       &                    -4.98\% &                    -9.54\% &                   -5.15\% &                    -9.70\% \\
Interior equipment [MWh]    &                   -17.92\% &                   -23.66\% &                  -15.40\% &                   -21.24\% \\
Fans [MWh]                  &                   -15.41\% &                   -22.48\% &                  -17.49\% &                   -23.84\% \\
Pumps [MWh]                 &                   -13.65\% &                   -18.52\% &                  -17.19\% &                   -22.59\% \\
Water heating, Gas [MWh]    &                    -9.02\% &                   -13.93\% &                   -9.27\% &                   -14.67\% \\
PV Generation [MWh]         &                   -37.29\% &                   -45.21\% &                  -31.79\% &                   -41.56\% \\
Heating demand [MWh]        &                   -10.35\% &                   -13.89\% &                   -9.70\% &                   -13.68\% \\
Cooling demand [MWh]        &                    -5.33\% &                    -7.79\% &                   -5.42\% &                    -7.43\% \\
\bottomrule
\end{tabular}

\end{footnotesize}
\end{minipage}
\caption{\textbf{Recorded surrogate model error reduction after transferring uncertain samples to the high-fidelity simulation model.} The data shows the error if either 100\%, 90\% or 80\% of the building design samples are processed by the surrogate model. If 10\% or 20\% are processed by the high-fidelity simulation model, errors produced by the surrogate can be avoided and the overall error decreases (here quantified by the 90-percentile absolute percentage error).}
\label{fig:filt_heating}
\end{figure}

In this section we study how effectively the epistemic uncertainty estimates can be used to predict inaccuracies of the surrogate model.\\
The concept is as follows. We sort the uncertainty estimates on the test data by scale to identify samples where surrogate model estimates are inaccurate. Samples with high uncertainty will be simulated using the high fidelity simulation program instead of the surrogate model (see Figure \ref{fig:concept}). As a consequence the user of the surrogate model, here a building designer, is facing lower inaccuracies. This should be traded-off against an increase in runtime, as the expensive high-fidelity engineering simulation is run. This trade-off is handled by defining the uncertainty threshold above which the simulation program is queried.\\
We define this threshold as the 90- or 80-percentile of all uncertainties observed on our test data set. The rationale behind that choice is that only 10\% (or 20\%) of all samples are transferred to the slow simulation program to not increase computational cost heavily. In reality, finding a suitable threshold is more difficult and could also be based on project-related uncertainty requirements, e.g. accepted uncertainty in energy consumption.\\
In Figure \ref{fig:filt_heating}, the decrease in the error of the surrogate model predictions is illustrated for the three target variables covering the heat supply of different fuel sources. These targets produced the largest errors (see Section \ref{sec:surrogate_accuracy}) and thus, we focus on increasing the surrogate robustness particularly for them. Discarding the 10\% samples with the highest uncertainty on the test data, we can decrease the $APE_{90}$ error in estimating the annual heating supply with a gas furnace from 24.9\% to 18.9\%.\footnote{The 18.9\% error was computed on the 90\% remaining samples in the test set.} This is equivalent to a reduction of $\approx 25\%$.\\
The $MAPE$ error on the other surrogate model outputs was reduced by 4\% to 18\%, and the $APE_{90}$ by 5\% to 25\% (see Figure \ref{fig:filt_heating}). In particular, the significant reduction of the $APE_{90}$ error proofs the increase in the robustness of the surrogate model predictions.

%% file: 05DiscussionConclusion.tex
\section{Discussion}
\label{sec:discussion}
Surrogate models have shown to help architects and building designers to rapidly assess the energy performance of their designs \cite{brown2020}. However, by being only approximative, concerns about the robustness of the surrogate model accuracy arise. A Bayesian approach for surrogate modelling, allows to not only provide a performance estimate but also inform about the confidence of the approximating surrogate model and potentially, to identify parts of the design space where the surrogate model may provide inaccurate results. \\
This first analysis of the use of Bayesian surrogate models revealed essential properties on the robustness of surrogate models, and how Bayesian modelling can be an aid for effective reasoning on the energy performance of buildings under the epistemic uncertainty of surrogates. The goal was to augment surrogates such that we can maintain the benefits of surrogate models while minimizing the risk associated with the uncertainty of surrogate models.

\subsection{Lacking robustness of surrogate models} 
Surrogate model accuracy is often reported with error metrics like the $R^2$ or $MAPE$ score. They are important but can be deceiving. A high coefficient of explained variance ($R^2$) or a low mean absolute percentage error $MAPE$, may disguise that the surrogate may produce quite large errors in certain fractions of the design space. For example, we found that the 90-percentile absolute percentage error can be as high as $22.3\%$ although an $R^2=0.99$ suggests very high performance (see Table \ref{tab:first}). This motivates, that indeed measures to identify surrogate inaccuracies could lessen the risk associated with surrogate modelling.

\subsection{Bayesian learning to express surrogate confidence}
Results on the quality of uncertainty estimates of the dropout neural network validated that it can be used to effectively express confidence on its predictions, e.g. one can formulate that the heating demand for a building with a wall of 1$m$ thickness is between $220 MWh/year$ and $230 MWh/year$ with a 90\% confidence (see Figure \ref{fig:example_functionfamily}).\\
On the other hand, while being almost as accurate as the neural network model, we found that the stochastic variational Gaussian Process model produces miscalibrated uncertainty estimates. Please note, that this finding cannot be generalized as methods exist to calibrate uncalibrated estimates \cite{kuleshov2018accurate}, and in other studies deep Gaussian process models were found to produce a larger variance in the uncertainty estimates \cite{salimbeni2017doubly}. Nonetheless, the results on the SVGP models highlight that assessing the quality of Bayesian uncertainty estimates is important.

\subsection{Bayesian learning to identify erroneous surrogate estimates}
\begin{figure}
\includegraphics[scale=.6]{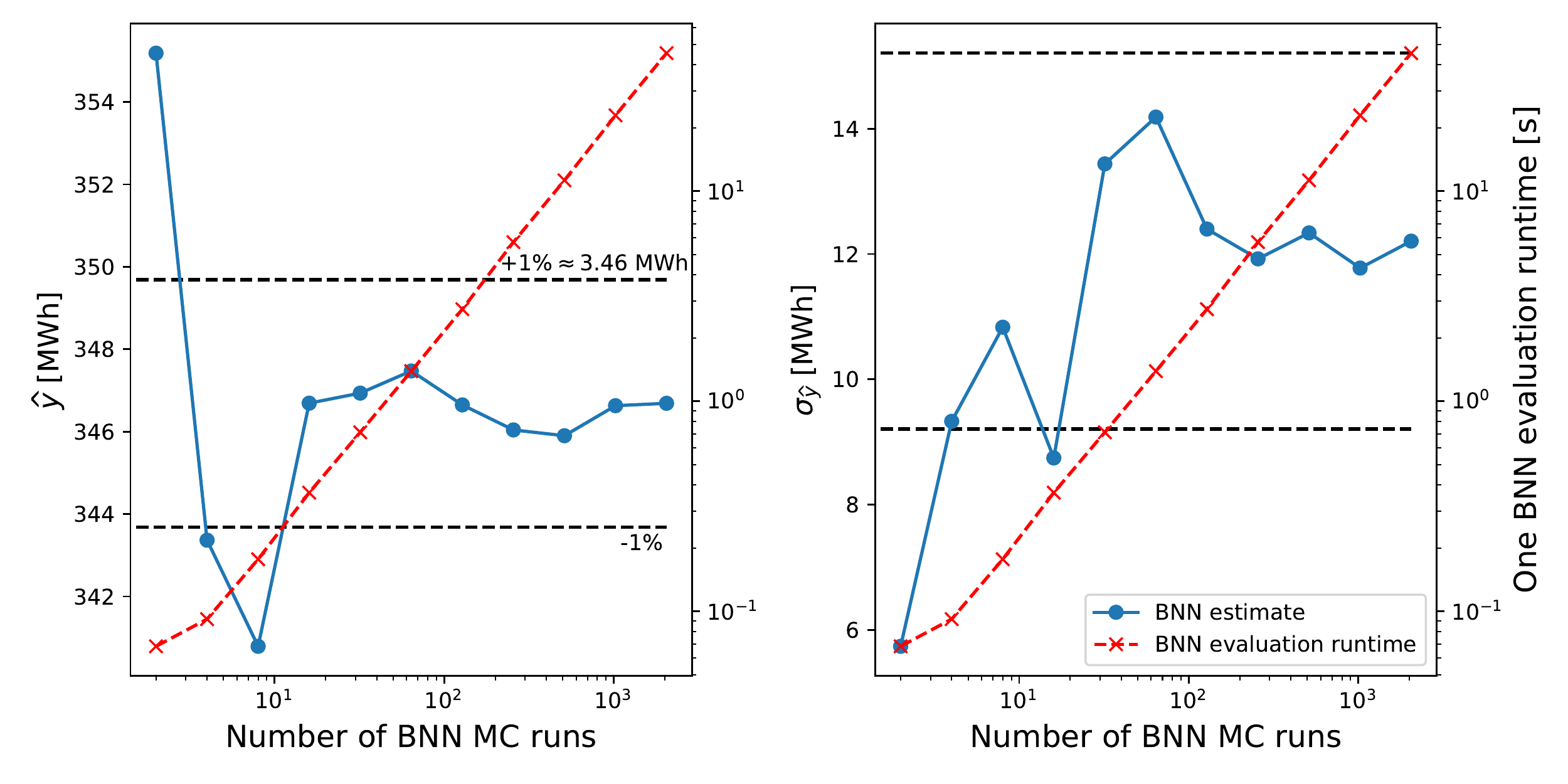}
\caption{\textbf{Convergence of BNN estimates with an increasing number of Monte Carlo dropout samples.} The plot shows BNN heating demand estimates and uncertainty estimates with increasing number of MC samples (see case study in Section \ref{sec:case_study}). Both approximately converge after conducting 30 random dropout runs, which takes around 0.8 seconds (without parallelization). }
\label{fig:convergence}
\end{figure}

We leveraged the uncertainty estimates to express warnings when the surrogate model is highly uncertain. By defining a threshold, here the 90-percentile or 80-percentile of the uncertainty estimates on the test data, we could reduce the $APE_{90}$ error by up to 40\%.\\
This is a significant first step towards the hybridization of fast, low-fidelity and slow, high-fidelity models. Still, practical issues have to be solved. For example, the question arises on how to implement the high-fidelity model runs. They could be carried out in the background while the surrogate model user would be working with the vague estimates as a start. In our case the results would be updated after 2 minutes and 10 seconds, which corresponds to the approximate runtime of one simulation.\\
Another issue is that the computational cost of evaluating a Bayesian model increases compared to a deterministic surrogate model. This is particularly the case for BNNs, whose uncertainty estimates are generated with Monte Carlo (MC) dropout. The BNN estimates converge with an increasing numbers of BNN MC evaluations, which is shown in Figure \ref{fig:convergence}. The plot implies that uncertainty estimates for a single sample take approximately 0.8 seconds to guarantee convergence. This may be too slow for interactive engineering design tasks but can be easily fixed parallelizing the MC dropout sampling.\\
These and other questions have to be addressed when integrating Bayesian surrogates into software products for building designers.

\subsection{Accuracy of the Bayesian model compared to a deterministic surrogate model}
We can compare the results of this study to a non-Bayesian feed-forward neural network trained on the same dataset (see Table \ref{tab:comparison} in the Appendix). Details on the non-bayesian network used can be found in \cite{PaulWestermann2020}. It has a very similar layout to the dropout BNN (2 hidden layers with 512 neurons, leaky rectified linear unit activation function) and was trained using the same cost function and optimizer (1200 training epochs with Adam optimizer).\\
The $R^2$, $MAPE$ and $APE_90$ scores of the deterministic model computed on the test data are better for most outputs when no uncertainty based sample filtering is applied (see Table \ref{tab:comparison}). However, when using uncertainty thresholds the Bayesian model produces lower $MAPE$ and $APE_90$ errors proposing that the BNN is a useful means to increase robustness of surrogate models.\footnote{Here, we used a uniformly distributed set of building design samples as our test data. However, this may not be representative of actual design processes. In future, a comparison of both neural network types (Baysian surrogate model, non-bayesian surrogate model) that takes architectural design preferences into account when constructing a set of test data should be considered.}\\

\section{Conclusion and Outlook}
In this study we proposed to augment and hybridize physics-based simulation software with Bayesian (deep) learning surrogate models. By quantifying the surrogate model (epistemic) uncertainty, the Bayesian paradigm acknowledges that surrogate models are approximations of original simulation models, and it offers a tool to effectively reason under that incurred uncertainty while exploiting the much faster runtime of surrogate models to produce engineering performance estimates.\\
In a case study we showcased the application of Bayesian surrogate models for the design of net-zero energy buildings. We found that dropout neural network models provided well-calibrated uncertainty estimates, which can be used to identify building design choices for which the surrogate, that estimates the associated building energy performance, produces large errors. The latter enables us to refer those designs to the high-fidelity energy simulation tool to assure accurate estimates for the architect or building designer. That referral process significantly lowered the errors in comparison to a common deterministic surrogate model.\\
Although all findings are bound to the case study of a building simulation surrogate, results motivate to apply Bayesian learning to other fields where surrogate models are common.\\

In future, we foresee that Bayesian models will allow us to further \textit{hybridize} data-driven surrogate models and high-fidelity simulation models \cite{reichstein2019deep}. For that purpose enriching the Bayesian surrogate models with physical know-how could be a key element. Furthermore, Bayesian learning forms a foundation for adaptively sampling simulation runs, for which the surrogate model is particularly uncertain. This progress, called active learning, will be explored in an upcoming study \cite{inproceedings}.

\section*{Code and Data availability}
The entire source code of this work, the EnergyPlus description file (\textit{.idf}) of the building template, and instructions on how to download the data used in this study are available in a GitLab repository.\footnote{\url{https://gitlab.com/energyincities/building_surrogate_modelling}}

\section*{Acknowledgements}
\label{sec:ack}
This research was supported by grant funding from CANARIE via the BESOS project (CANARIE RS-327). 

%% file: main.bbl
\begin{thebibliography}{10}
\expandafter\ifx\csname url\endcsname\relax
  \def\url#1{\texttt{#1}}\fi
\expandafter\ifx\csname urlprefix\endcsname\relax\def\urlprefix{URL }\fi
\expandafter\ifx\csname href\endcsname\relax
  \def\href#1#2{#2} \def\path#1{#1}\fi

\bibitem{iea_2019}
C.~D. John~Dulac, Thibaut~Abergel,
  \href{https://www.iea.org/reports/tracking-buildings}{Tracking buildings},
  Tech. rep., {Internation Energy Agency} (2019).
\newline\urlprefix\url{https://www.iea.org/reports/tracking-buildings}

\bibitem{Westermann2019}
P.~Westermann, R.~Evins, Surrogate modelling for sustainable building design
  {\textendash} a review, Energy and Buildings 198 (2019) 170--186.
\newblock \href {https://doi.org/10.1016/j.enbuild.2019.05.057}
  {\path{doi:10.1016/j.enbuild.2019.05.057}}.

\bibitem{jusselme_2020}
T.~Jusselme, Data-driven method for low-carbon building design at early stages,
  Ph.D. thesis, EPF Lausanne (2020).

\bibitem{bpf}
{Open Technologies}, \href{http://www.buildingpathfinder.com/}{The building
  pathfinder}, Online.
\newline\urlprefix\url{http://www.buildingpathfinder.com/}

\bibitem{PaulWestermann2020}
{Paul Westermann}, {David Rulff}, {Kevin Cant}, {Gaelle Faure}, {Ralph Evins},
  \href{http://www.enerarxiv.org/page/thesis.html?id=1975}{Net-zero navigator:
  A platform for interactive net-zero building design using surrogate
  modelling}.
\newline\urlprefix\url{http://www.enerarxiv.org/page/thesis.html?id=1975}

\bibitem{waibel2019building}
C.~Waibel, T.~Wortmann, R.~Evins, J.~Carmeliet, Building energy optimization:
  An extensive benchmark of global search algorithms, Energy and Buildings 187
  (2019) 218--240.

\bibitem{rivalin2018comparison}
L.~Rivalin, P.~Stabat, D.~Marchio, M.~Caciolo, F.~Hopquin, A comparison of
  methods for uncertainty and sensitivity analysis applied to the energy
  performance of new commercial buildings, Energy and Buildings 166 (2018)
  489--504.

\bibitem{RN363}
J.~Hester, J.~Gregory, R.~Kirchain, \href{<Go to
  ISI>://WOS:000390624800018}{Sequential early-design guidance for residential
  single-family buildings using a probabilistic metamodel of energy
  consumption}, Energy and Buildings 134 (2017) 202--211.
\newblock \href {https://doi.org/10.1016/j.enbuild.2016.10.047}
  {\path{doi:10.1016/j.enbuild.2016.10.047}}.
\newline\urlprefix\url{<Go to ISI>://WOS:000390624800018}

\bibitem{brown2020}
N.~C. Brown, Design performance and designer preference in an interactive,
  data-driven conceptual building design scenario, Design Studies (2020) {}.

\bibitem{RN351}
T.~Ostergard, R.~L. Jensen, S.~E. Maagaard, \href{<Go to
  ISI>://WOS:000425075600008}{A comparison of six metamodeling techniques
  applied to building performance simulations}, Applied Energy 211 (2018)
  89--103.
\newblock \href {https://doi.org/10.1016/j.apenergy.2017.10.102}
  {\path{doi:10.1016/j.apenergy.2017.10.102}}.
\newline\urlprefix\url{<Go to ISI>://WOS:000425075600008}

\bibitem{westermann2020weather}
P.~Westermann, R.~Evins, Using a deep temporal convolutional network as a
  building energy surrogate model that spans multiple climate zones, Applied
  Energy 264 (2020) 114715.

\bibitem{kendall2017uncertainties}
A.~Kendall, Y.~Gal, What uncertainties do we need in bayesian deep learning for
  computer vision?, in: Advances in neural information processing systems,
  2017, pp. 5574--5584.

\bibitem{damianou2013deep}
A.~Damianou, N.~Lawrence, Deep gaussian processes, in: Artificial Intelligence
  and Statistics, 2013, pp. 207--215.

\bibitem{gal2016dropout}
Y.~Gal, Z.~Ghahramani, Dropout as a bayesian approximation: Representing model
  uncertainty in deep learning, in: International Conference on Machine
  Learning, 2016, pp. 1050--1059.

\bibitem{hensman2013gaussian}
J.~Hensman, N.~Fusi, N.~D. Lawrence, Gaussian processes for big data, in:
  Uncertainty in Artificial Intelligence, Citeseer, 2013, p. 282.

\bibitem{filos2019systematic}
A.~Filos, S.~Farquhar, A.~N. Gomez, T.~G. Rudner, Z.~Kenton, L.~Smith,
  M.~Alizadeh, A.~de~Kroon, Y.~Gal, A systematic comparison of bayesian deep
  learning robustness in diabetic retinopathy tasks, arXiv preprint
  arXiv:1912.10481.

\bibitem{reichstein2019deep}
M.~Reichstein, G.~Camps-Valls, B.~Stevens, M.~Jung, J.~Denzler, N.~Carvalhais,
  et~al., Deep learning and process understanding for data-driven earth system
  science, Nature 566~(7743) (2019) 195--204.

\bibitem{wang2007review}
G.~G. Wang, S.~Shan, Review of metamodeling techniques in support of
  engineering design optimization, Journal of Mechanical design 129~(4) (2007)
  370--380.

\bibitem{RN401}
F.~Ritter, P.~Geyer, A.~Borrmann, Simulation-based decision-making in early
  design stages, in: 32nd CIB W78 conference, Eindhoven, The Netherlands, 2015,
  pp. 27--29.

\bibitem{vazquez2019deep}
J.~Vazquez-Canteli, A.~D. Demir, J.~Brown, Z.~Nagy, Deep neural networks as
  surrogate models for urban energy simulations, in: Journal of Physics:
  Conference Series, Vol. 1343, IOP Publishing, 2019, p. 012002.

\bibitem{RN397}
A.~Prada, A.~Gasparella, P.~Baggio, On the performance of meta-models in
  building design optimization, Applied Energy 225 (2018) 814--826.

\bibitem{RN389}
B.~Eisenhower, Z.~O'Neill, S.~Narayanan, V.~A. Fonoberov, I.~Mezic, \href{<Go
  to ISI>://WOS:000301989800034}{A methodology for meta-model based
  optimization in building energy models}, Energy and Buildings 47 (2012)
  292--301.
\newblock \href {https://doi.org/10.1016/j.enbuild.2011.12.001}
  {\path{doi:10.1016/j.enbuild.2011.12.001}}.
\newline\urlprefix\url{<Go to ISI>://WOS:000301989800034}

\bibitem{bre2020efficient}
F.~Bre, N.~Roman, V.~D. Fachinotti, An efficient metamodel-based method to
  carry out multi-objective building performance optimizations, Energy and
  Buildings 206 (2020) 109576.

\bibitem{RN414}
C.~J. Hopfe, J.~L. Hensen, Uncertainty analysis in building performance
  simulation for design support, Energy and Buildings 43~(10) (2011)
  2798--2805.

\bibitem{coakley2014review}
D.~Coakley, P.~Raftery, M.~Keane, A review of methods to match building energy
  simulation models to measured data, Renewable and sustainable energy reviews
  37 (2014) 123--141.

\bibitem{RN383}
M.~Manfren, N.~Aste, R.~Moshksar, \href{<Go to
  ISI>://WOS:000314669500059}{Calibration and uncertainty analysis for computer
  models - a meta-model based approach for integrated building energy
  simulation}, Applied Energy 103 (2013) 627--641.
\newblock \href {https://doi.org/10.1016/j.apenergy.2012.10.031}
  {\path{doi:10.1016/j.apenergy.2012.10.031}}.
\newline\urlprefix\url{<Go to ISI>://WOS:000314669500059}

\bibitem{heo2012calibration}
Y.~Heo, R.~Choudhary, G.~Augenbroe, Calibration of building energy models for
  retrofit analysis under uncertainty, Energy and Buildings 47 (2012) 550--560.

\bibitem{sokol2017validation}
J.~Sokol, C.~C. Davila, C.~F. Reinhart, Validation of a bayesian-based method
  for defining residential archetypes in urban building energy models, Energy
  and Buildings 134 (2017) 11--24.

\bibitem{kristensen2018hierarchical}
M.~H. Kristensen, R.~E. Hedegaard, S.~Petersen, Hierarchical calibration of
  archetypes for urban building energy modeling, Energy and Buildings 175
  (2018) 219--234.

\bibitem{garud2017design}
S.~S. Garud, I.~A. Karimi, M.~Kraft, {D}esign of computer experiments: {A}
  review, Computers \& Chemical Engineering 106 (2017) 71--95.

\bibitem{crawley2001energyplus}
D.~B. Crawley, L.~K. Lawrie, F.~C. Winkelmann, W.~F. Buhl, Y.~J. Huang, C.~O.
  Pedersen, R.~K. Strand, R.~J. Liesen, D.~E. Fisher, M.~J. Witte, et~al.,
  Energyplus: creating a new-generation building energy simulation program,
  Energy and buildings 33~(4) (2001) 319--331.

\bibitem{rasmussen2004gaussian}
C.~E. Rasmussen, Gaussian processes in machine learning, in: Advanced lectures
  on machine learning, Springer, 2004, pp. 63--71.

\bibitem{ostergaard2016building}
T.~{\O}sterg{\aa}rd, R.~L. Jensen, S.~E. Maagaard,
  \href{https://www.sciencedirect.com/science/article/pii/S136403211600280X}{Building
  simulations supporting decision making in early design--a review}, Renewable
  and Sustainable Energy Reviews 61 (2016) 187--201.
\newline\urlprefix\url{https://www.sciencedirect.com/science/article/pii/S136403211600280X}

\bibitem{blei2017variational}
D.~M. Blei, A.~Kucukelbir, J.~D. McAuliffe, Variational inference: A review for
  statisticians, Journal of the American statistical Association 112~(518)
  (2017) 859--877.

\bibitem{neal1995bayesian}
R.~M. Neal, Bayesian learning for neural networks, Vol. 118, Springer Science
  \& Business Media, 1995.

\bibitem{gal2016uncertainty}
Y.~Gal, Uncertainty in deep learning, University of Cambridge 1~(3).

\bibitem{pearce2018uncertainty}
T.~Pearce, M.~Zaki, A.~Brintrup, N.~Anastassacos, A.~Neely, Uncertainty in
  neural networks: Bayesian ensembling, arXiv preprint arXiv:1810.05546.

\bibitem{srivastava2014dropout}
N.~Srivastava, G.~Hinton, A.~Krizhevsky, I.~Sutskever, R.~Salakhutdinov,
  Dropout: a simple way to prevent neural networks from overfitting, The
  journal of machine learning research 15~(1) (2014) 1929--1958.

\bibitem{chollet2015keras}
F.~Chollet, et~al., Keras (2015).

\bibitem{abadi2016tensorflow}
M.~Abadi, P.~Barham, J.~Chen, Z.~Chen, A.~Davis, J.~Dean, M.~Devin,
  S.~Ghemawat, G.~Irving, M.~Isard, et~al., Tensorflow: a system for
  large-scale machine learning., in: OSDI, Vol.~16, 2016, pp. 265--283.

\bibitem{titsias2009variational}
M.~Titsias, Variational learning of inducing variables in sparse gaussian
  processes, in: Artificial Intelligence and Statistics, 2009, pp. 567--574.

\bibitem{bauer2016understanding}
M.~Bauer, M.~van~der Wilk, C.~E. Rasmussen, Understanding probabilistic sparse
  gaussian process approximations, in: Advances in neural information
  processing systems, 2016, pp. 1533--1541.

\bibitem{salimbeni2017doubly}
H.~Salimbeni, M.~Deisenroth, Doubly stochastic variational inference for deep
  gaussian processes, in: Advances in Neural Information Processing Systems,
  2017, pp. 4588--4599.

\bibitem{svendsen2020deep}
D.~H. Svendsen, P.~Morales-{\'A}lvarez, A.~B. Ruescas, R.~Molina,
  G.~Camps-Valls, Deep gaussian processes for biogeophysical parameter
  retrieval and model inversion, ISPRS Journal of Photogrammetry and Remote
  Sensing 166 (2020) 68--81.

\bibitem{gpy2014}
{GPy}, {GPy}: A gaussian process framework in python,
  \url{http://github.com/SheffieldML/GPy} (since 2012).

\bibitem{crawley2000energyplus}
D.~B. Crawley, C.~O. Pedersen, L.~K. Lawrie, F.~C. Winkelmann, Energyplus:
  energy simulation program, ASHRAE journal 42~(4) (2000) 49.

\bibitem{NECB}
National Research Council Canada (NRCan),
  \href{https://nrc.canada.ca/en/certifications-evaluations-standards/codes-canada/codes-canada-publications/national-energy-code-canada-buildings-2017}{National
  Energy Code of Canada for Buildings 2017} (2017).
\newline\urlprefix\url{https://nrc.canada.ca/en/certifications-evaluations-standards/codes-canada/codes-canada-publications/national-energy-code-canada-buildings-2017}

\bibitem{box1964analysis}
G.~E. Box, D.~R. Cox, An analysis of transformations, Journal of the Royal
  Statistical Society: Series B (Methodological) 26~(2) (1964) 211--243.

\bibitem{RN358}
R.~E. Edwards, J.~New, L.~E. Parker, B.~Cui, J.~Dong, \href{<Go to
  ISI>://WOS:000407188500055}{Constructing large scale surrogate models from
  big data and artificial intelligence}, Applied Energy 202 (2017) 685--699.
\newblock \href {https://doi.org/10.1016/j.apenergy.2017.05.155}
  {\path{doi:10.1016/j.apenergy.2017.05.155}}.
\newline\urlprefix\url{<Go to ISI>://WOS:000407188500055}

\bibitem{RN399}
A.~Rackes, A.~P. Melo, R.~Lamberts, \href{<Go to
  ISI>://WOS:000377728700022}{Naturally comfortable and sustainable: Informed
  design guidance and performance labeling for passive commercial buildings in
  hot climates}, Applied Energy 174 (2016) 256--274.
\newblock \href {https://doi.org/10.1016/j.apenergy.2016.04.081}
  {\path{doi:10.1016/j.apenergy.2016.04.081}}.
\newline\urlprefix\url{<Go to ISI>://WOS:000377728700022}

\bibitem{kuleshov2018accurate}
V.~Kuleshov, N.~Fenner, S.~Ermon, Accurate uncertainties for deep learning
  using calibrated regression, in: International Conference on Machine
  Learning, 2018, pp. 2796--2804.

\bibitem{platt1999probabilistic}
J.~Platt, et~al., Probabilistic outputs for support vector machines and
  comparisons to regularized likelihood methods, Advances in large margin
  classifiers 10~(3) (1999) 61--74.

\bibitem{scalia2020evaluating}
G.~Scalia, C.~A. Grambow, B.~Pernici, Y.-P. Li, W.~H. Green, Evaluating
  scalable uncertainty estimation methods for deep learning based molecular
  property prediction, Journal of Chemical Information and Modeling.

\bibitem{inproceedings}
P.~Westermann, R.~Evins, Adaptive sampling for building simulation surrogate
  model derivation using the lola-voronoi algorithm, in: {International
  Building Performance Association (IBPSA)} (Ed.), Proceedings of the
  International Building Performance Simulation Association, Vol.~16, 2019, pp.
  1559--1563.
\newblock \href {https://doi.org/10.26868/25222708.2019.211232}
  {\path{doi:10.26868/25222708.2019.211232}}.

\end{thebibliography}
